\documentclass{article}

\usepackage[eandd,final]{neurips_2026}
\usepackage[colorlinks=true,
            citecolor=black,
            linkcolor=black,
            urlcolor=black]{hyperref}

\usepackage[utf8]{inputenc} 
\usepackage[T1]{fontenc}    
\usepackage{hyperref}       
\usepackage{url}            
\usepackage{booktabs}       
\usepackage{amsfonts}       
\usepackage{nicefrac}       
\usepackage{microtype}      
\usepackage{xcolor}         

\usepackage{graphicx}
\usepackage{subcaption}
\usepackage{placeins}

\usepackage{amsmath,amsfonts,bm}

\def\eqref#1{equation~\ref{#1}}

\def\1{\bm{1}}

\DeclareMathAlphabet{\mathsfit}{\encodingdefault}{\sfdefault}{m}{sl}
\SetMathAlphabet{\mathsfit}{bold}{\encodingdefault}{\sfdefault}{bx}{n}

\usepackage{hyperref}
\usepackage{url}

\title{PMO-Dock: Benchmarking Docking, Specificity, and Generalization in Molecular Optimization}

\author{
Gor Simonyan\thanks{Denotes equal first authorship. Correspondence to Gor Simonyan, Tatevik Abrahamyan (\texttt{gor@yerevann.com}, \texttt{tatevikabrahamyan@yerevann.com}).}\\
YerevaNN\\
Yerevan State University
\And
Tatevik Abrahamyan\footnotemark[1]\\
YerevaNN\\
Yerevan State University
\And
Narek Abelyan\\
YerevaNN\\
Russian-Armenian University\\
\And
Tigran Fahradyan\\
University of Cambridge
\And
Hrant Khachatrian\\
YerevaNN\\
Yerevan State University
}

\begin{document}

\maketitle

\begin{abstract}
The Practical Molecular Optimization (PMO) benchmark standardized evaluation in molecular optimization, but it is built on simple property-based oracles that do not capture structure-based drug design. As the field has moved to docking-based objectives, methods are often evaluated on different docking tasks, without a shared protocol or a shared limit on oracle calls. To address this, we introduce \textbf{P}ractical \textbf{M}olecular \textbf{O}ptimization for \textbf{Dock}ing (PMO-Dock), a benchmark and protocol for docking-based optimization consisting of 25 tasks covering hit generation, lead optimization, and new specificity tasks that require strong on-target binding while penalizing off-target binding. The protocol separates model development from final evaluation: validation tasks are used for hyperparameter tuning, and test tasks are held out for final evaluation and are not used during development. Evaluation on the test tasks is done under a strict oracle budget to reflect realistic drug-discovery settings. We benchmark four methods from different optimization paradigms: Saturn (reinforcement learning), GenMol (discrete diffusion), Genetic-guided GFlowNet, and ChemLlama (language-model evolutionary search). No method is best on every task. Hyperparameter transfer across tasks and across categories is often weak, and it differs by method, so hyperparameters are chosen within each task category on the validation tasks and then frozen for the test tasks. The benchmark measures generalization to unseen tasks and sample efficiency under a limited test-time oracle budget.
\end{abstract}

\begin{figure*}[ht]
\centering
\includegraphics[width=0.75\linewidth]{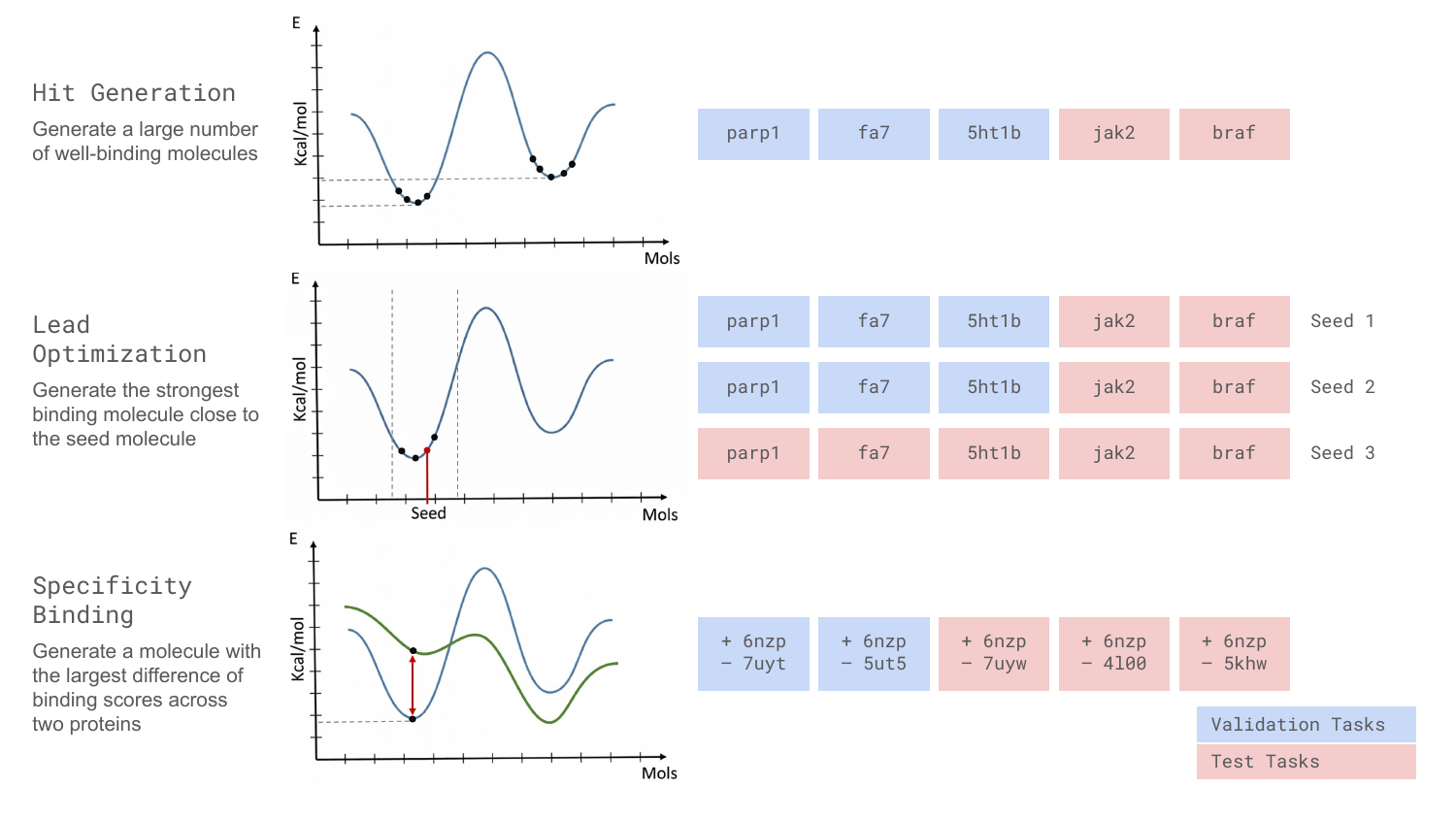}
\caption{Overview of all 25 PMO-Dock tasks organized by oracle category. Each task is shown as a separate cell, color-coded by split: validation tasks (light blue) are used for algorithm development and hyperparameter tuning, while test tasks (light red) are strictly held out for evaluation with limited evaluation budgets. Hit Generation tasks maximize the fraction of molecules meeting docking thresholds, Lead Optimization tasks optimize docking scores while maintaining similarity to seed molecules and specificity tasks maximize the gap between on-target and off-target docking scores.}
\label{fig:all_tasks}
\end{figure*}

\section{Introduction and Related Work}

Small-molecule optimization requires search over a large chemical space under objectives that are often expensive to evaluate. Many methods have been proposed, but they are evaluated on different tasks and protocols, so results are difficult to compare. 

The Practical Molecular Optimization (PMO) benchmark~\citep{gao2022sample} addressed this by establishing rigorous evaluation standards. PMO compiled 23 optimization tasks with well-specified oracles and evaluated 23 methods using unified protocols: strict oracle budgets, extensive hyperparameter sweeps, and fixed metrics. That shared protocol made later comparisons possible. Methods such as Augmented Memory~\citep{guo2023augmented}, Genetic-guided GFN~\citep{kim2024genetic}, Chemlactica~\citep{guevorguian2024chemlactica}, and MOLLEO~\citep{wang2024molleo} then advanced the state of the art on PMO. PMO's objectives, however, combine simple RDKit properties, such as molecular similarity, isomer counts, and drug-likeness. They do not include docking against a protein target.

Recently, an increasing number of methods adopt molecular docking objectives. Methods including Augmented Memory~\citep{guo2023augmented}, Beam Enumeration~\citep{guo2023beam}, Saturn~\citep{guo2024saturn}, GenMol~\citep{genmol}, and Chemlactica~\citep{guevorguian2024chemlactica} have explored docking score optimization. \citet{cieplinski2023dock} introduced a benchmark based on SMINA docking for de novo design, in which a model trained on known ligands for a target proposes a new batch of molecules scored on that same target. That benchmark measures only single-target hit optimization, and it guides search with a learned surrogate rather than a live docking oracle. TARTARUS~\citep{nigam2023tartarus} evaluates generated molecules with a live docking oracle during optimization. Both benchmarks optimize individual tasks. Neither holds tasks out to test whether a model and its hyperparameters generalize to unseen proteins or to a different design setting. PMO-Dock pairs docking objectives with a PMO-style split between development and final evaluation.


Our contributions are as follows: 
\begin{enumerate}
    \item We introduce PMO-Dock, a benchmark that applies a PMO-style protocol to docking objectives. It has 25 tasks, 11 for development and 14 held out for final evaluation, covering hit generation, lead optimization, and specificity.
    
    \item We add a specificity category in which a molecule must bind a chosen pocket and bind a designated off-target pocket less strongly. The category has five tasks, covering pockets of the same protein and pockets of different proteins.

    \item We evaluate four molecular optimization algorithms. None is best on every task: Saturn leads on hit generation, and ChemLlama leads on specificity. We also measure how far hyperparameters transfer across tasks, and how much performance drops when the task-best configuration is not used. Configurations tuned on similar hit-generation targets can be reused for Saturn. When transfer is weak, the test task needs a validation task of the same category.
\end{enumerate}

\section{The PMO-Dock Benchmark}\label{sec:benchmark}

This section defines the PMO-Dock benchmark. It measures whether a method can find high-scoring molecules in a small number of oracle calls, and whether that behavior carries over to tasks it was not tuned on. The benchmark has three categories, \textbf{hit generation}, \textbf{lead optimization}, and \textbf{specificity}, each with its own success metric.


\subsection{Hit Generation}

We use the hit-generation task from Saturn~\citep{guo2024saturn}. A molecular generative model proposes candidate molecules for a given protein target and is evaluated by an oracle under a fixed evaluation budget. The optimization objective is to maximize the number of generated unique molecules that satisfy predefined physicochemical and target-specific constraints.\\\textbf{Targets.}~The benchmark includes five protein targets:
PARP1, Factor VII (FA7), the 5-hydroxytryptamine receptor 1B (5HT1B), BRAF, and JAK2.\\\textbf{Optimization objective.}~Given an oracle budget of $B = 3000$ molecule evaluations, a generative policy $\pi$ induces a set of evaluated molecules $\{x_i\}_{i=1}^B$. Let $\mathcal{H}(x) \in \{0, 1\}$ denote a binary indicator of whether molecule $x$ satisfies the hit constraints. The objective is to
maximize the expected \textbf{hit rate},
\[
   \max_{\pi} \; \mathbb{E}_{\{x_i\}_{i=1}^B \sim \pi} \left[\frac{1}{B}\sum_{i=1}^{B} \mathcal{H}(x_i)\right],
\]
We run each algorithm with ten random seeds and report the average metric.\\\textbf{Hit definition.}
\[
\mathcal{H}(x_i) =
\begin{cases}
1, & \text{if } \mathrm{QED}(x)\in[0.5,1.0]\ \land\ \mathrm{SA}(x)\in[1.0,5.0]\ \land\ s^{(t)}(x)\ge s_{\min}^{(t)},\\
0, & \text{otherwise.}
\end{cases}
\]
where $s^{(t)}(x)$ denotes the docking value for target $t$, taken in absolute value. The target-specific lower bounds are given in Appendix \ref{app:task-thresholds}. Following prior work, we calculate $s^{(t)}$ using QuickVina~\citep{alhossary2015quickvina} with low exhaustiveness ($1$) to save computational resources.
\subsection{Lead Optimization}

We consider a \emph{lead optimization} task, borrowed from GenMol~\citep{genmol}, in which a molecular generative model proposes new compounds that improve a protein-specific objective while remaining in the local chemical neighborhood of a given seed molecule for the target protein. The task is constrained generation under a fixed oracle budget. The protein targets are the same as in the hit generation task.\\\textbf{Optimization objective.}~Given a seed molecule $x_0$ and a target $t$, the objective is to generate molecules that maximize predicted target activity as measured by a docking-based oracle $s^{(t)}(x)$, while satisfying locality and drug-likeness constraints. As in hit generation, $s^{(t)}(x)$ is a QuickVina~\citep{alhossary2015quickvina} score computed at exhaustiveness $1$. For a generative policy $\pi$ and an oracle budget of $B=1000$ evaluations, the optimization problem is
\[
    \max_{\pi} \; \mathbb{E}_{\{x_i\}_{i=1}^B \sim \pi}\left[ \max_{1 \le i \le B} s^{(t)}(x_i)\mathcal{H}(x_i) \right],
\]
subject to all generated molecules $x_i$ satisfying the constraints defined below. 
\[
\mathcal{H}(x_i) =
\begin{cases}
1, & \text{if } \mathrm{QED}(x)\in[0.6,1.0]\ \land\ \mathrm{SA}(x)\in[1.0,4.0]\ \land\ s^{(t)}(x)\ge s_{\min}^{(t)}\ \land\ \mathrm{sim}(x,x_0)\ge 0.6,\\
0, & \text{otherwise.}
\end{cases}
\]
The reported score is the highest docking score among generated molecules that satisfy every constraint in $\mathcal{H}$. The seed is treated as known to the algorithm. If no generated molecule satisfies the constraints, or if the best constrained score is worse than the seed's docking score, we report the seed's score. We run each algorithm with three random seeds and report the average metric.

\subsection{Specificity Binding}

Specificity tasks test selective binding, both between proteins and between pockets of the same protein. The optimizer must propose molecules that bind the chosen pocket and bind a designated off-target pocket less strongly. Given an oracle budget of $B = 3000$ molecule evaluations, the five tasks use the same held-out split as the rest of the benchmark. To the best of our knowledge, our benchmark is the first one that incorporates such a selectivity task, which is a crucial problem in real-world drug discovery.

The specificity objective was defined as a difference in docking scores:
\[
    \max_{\pi} \; \mathbb{E}_{\{x_i\}_{i=1}^B \sim \pi}\left[ \max_{1 \le i \le B} \mathcal{L}(x_i)\mathcal{H}(x_i) \right],
\]
where
\[
\mathcal{L} = s^{(P_{\text{target}})}(x) - s^{(P_{\text{off}})}(x)
\]

and 
\[
\mathcal{H}(x_i) =
\begin{cases}
1, & \text{if } \mathrm{QED}(x)\in[0.4,1.0]\ \land\ \mathrm{SA}(x)\in[1.0,4.0]\ \land\ s^{(6\mathrm{NZP})}(x)\ge s_{\mathrm{ref}}^{(\mathrm{spec})}\\
0, & \text{otherwise.}
\end{cases}
\]
Here $P_{\text{target}}$ is the on-target pocket and $P_{\text{off}}$ is either a homologous protein or another domain of the same protein. $s^{(t)}(x)$ is the docking score for pocket $t$. The objective rewards a larger gap between the on-target and off-target scores. We run each algorithm with three random seeds and report the average metric.\\\textbf{Targets.}~To assess both inter-protein and intra-protein pocket specificity, we selected three closely related kinases from the JAK family: JAK1, TYK2, and JAK2, which share high sequence and structural similarity. Additionally, each of these proteins contains two domains: the active kinase domain (JH1, ATP-binding pocket) and the pseudokinase domain (JH2, allosteric pocket). These domains also exhibit some level of similarity but have subtle differences that may influence binding. The two domains, repeated across these three kinases, provide both inter-protein and intra-protein pairs. The structures used are JAK1 JH1 (PDB ID: 5KHW), JAK1 JH2 (PDB ID: 4L00), TYK2 JH1 (PDB ID: 7UYT), TYK2 JH2 (PDB ID: 6NZP), JAK2 JH1 (PDB ID: 7UYW), and JAK2 JH2 (PDB ID: 5UT5). From these structures we define five pairwise tasks in three groups. Every task uses TYK2 JH2 as the on-target pocket.

\begin{enumerate}
    \item \textbf{Intra-protein domain specificity (one task):} strong binding to TYK2 JH2 and weaker binding to TYK2 JH1.
    \item \textbf{Inter-protein pseudokinase specificity (two tasks):} strong binding to TYK2 JH2 and weaker binding to JAK1 JH2, or to JAK2 JH2.
    \item \textbf{Cross-domain inter-protein specificity (two tasks):} strong binding to TYK2 JH2 and weaker binding to JAK1 JH1, or to JAK2 JH1.
\end{enumerate}

Docking scores are QuickVina~\citep{alhossary2015quickvina} scores from three independent runs at exhaustiveness $8$. The docking procedure is detailed in Appendix~\ref{sec:docking-details}.\\\textbf{Constraints.}~To define task satisfaction criteria, we used Deucravacitinib (BMS-986165) - a first-in-class, highly selective TYK2 pseudokinase inhibitor approved for the treatment of moderate-to-severe plaque psoriasis - as a reference compound for TYK2 JH2 binding. The docking score of Deucravacitinib against TYK2 JH2 was used as a baseline reference. For all specificity tasks, a generated molecule was considered successful only if its docking score to TYK2 JH2 exceeded that of the reference compound, thereby ensuring strong target engagement.

\subsection{Validation and Test Split}
\label{sec:val-test-protocol}
To measure generalization to unseen targets, and to unseen seed molecules in lead optimization, we split each category into \textbf{validation tasks} and \textbf{test tasks}. Validation tasks are used for algorithm development and hyperparameter selection. Test tasks are used only for the final evaluation. Figure~\ref{fig:all_tasks} shows the three categories: 11 validation tasks and 14 test tasks.

\subsection{Model Development and Hyperparameter Tuning}
Algorithm developers may run as many hyperparameter sweeps, and explore as many model design choices, as they need on the validation tasks. Test tasks should be used only for the final evaluation, after model development and hyperparameter selection are complete. Importantly, nothing should be assumed about the specific proteins, lead molecules, or binding pockets in the test set when developing and tuning new algorithms. For instance, if hyperparameters are selected based on protein similarity to validation targets, similarity metrics must be defined and tuned exclusively on validation tasks, to ensure no information leakage. The benchmark does not require a particular training dataset. Any dataset is allowed, provided it is chosen without using the test proteins, lead molecules, or binding pockets.

\begin{figure}[t]
\centering
\includegraphics[width=\linewidth]{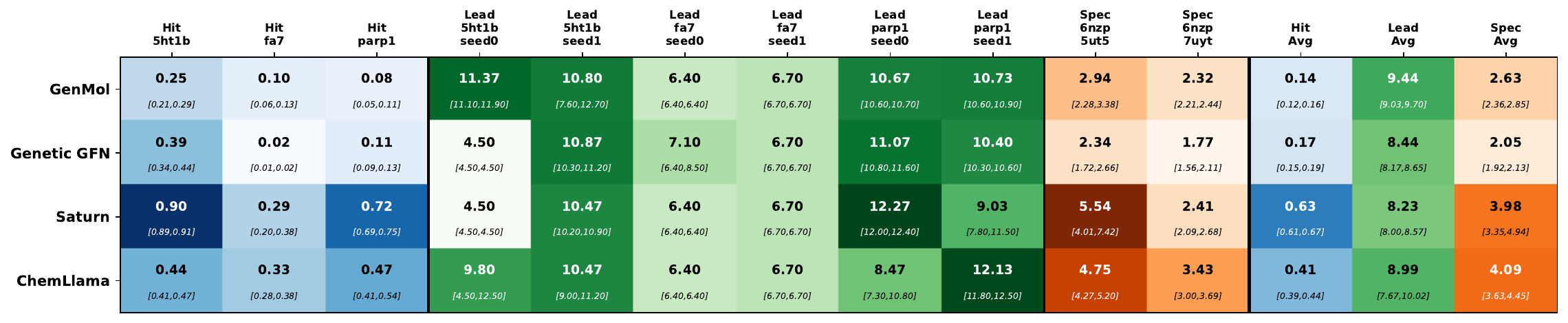}
\caption{Validation performance under category-specific hyperparameter selection. For each task, hyperparameters are chosen to maximize average performance on the other validation tasks in the same category (hit generation, lead optimization, or specificity). The last three columns are means over the tasks in each category (Hit Avg, Lead Avg, Spec Avg). Colors are normalized separately by category, so intensities should only be compared within category. More intense colors indicate higher performance relative to other entries in the same category. The alternative all-but-me rule is in Appendix~\ref{app:hparam-ablation}.}
\label{fig:perf_table}
\end{figure}

\section{Experiments}\label{sec:experiments}
In this section we tune and evaluate four recent molecular optimization algorithms on PMO-Dock.

\subsection{Optimization Methods}

We evaluate four diverse methods spanning different optimization paradigms: Saturn~\citep{guo2024saturn} (reinforcement learning with language models), GenMol~\citep{genmol} (discrete diffusion), Genetic-guided GFN~\citep{kim2024genetic} (GFlowNet with genetic operators), and ChemLlama (LLM-based evolutionary optimization). Saturn fine-tunes an autoregressive SMILES language model with reinforcement learning and replays high-reward molecules from an augmented memory buffer. GenMol is a discrete diffusion model over molecular fragments and optimizes by remasking and resampling fragments. Genetic-guided GFN fine-tunes a pretrained SMILES policy with the trajectory-balance objective while using genetic crossover and mutation to propose candidates. ChemLlama is our modification of Chemlactica~\citep{guevorguian2024chemlactica}. Chemlactica trains a Galactica language model on PubChem molecules and uses that model inside an evolutionary search. ChemLlama keeps the search and replaces the Galactica backbone with Llama. It maintains a pool of high-scoring molecules and prompts the model to propose analogs. Detailed descriptions of each method and their hyperparameter configurations are in Appendix~\ref{app:methods}.


\subsection{Hyperparameter Tuning}

Hyperparameters are chosen on the validation tasks only and then frozen for the test tasks. We use \textbf{category-specific selection}. Each method is swept independently on every validation task. The search spaces differ by method and are listed in Appendix~\ref{app:methods}. Saturn varies buffer size, reward type, and whether genetic operators are used. GenMol varies the fragment-masking rate, the randomness parameter, and the softmax temperature. Genetic-guided GFN varies the rank-based replay coefficient and the KL regularization coefficient. ChemLlama varies pool size, the number of prompts, and the number of similar molecules shown in each prompt. Hit rates, lead scores, and specificity scores are not on a common scale, so a configuration is never chosen by averaging across categories. For a validation task, each configuration is scored by its mean performance on the other validation tasks in the same category, and the configuration with the highest mean is selected. That configuration is then used for every test task in the category. Test scores are not used in the selection. An alternative that averages over all other validation tasks, regardless of category, is reported only as an ablation in Appendix~\ref{app:hparam-ablation}. Selecting the best configuration on each validation task itself, with no cross-task transfer, is a separate extended-budget ceiling, reported in Section~\ref{sec:extended-budget}.


\subsection{Validation Tasks}
\label{sec:validation-tasks}
Figure~\ref{fig:perf_table} reports validation performance under that category-specific rule, one row per method. Saturn is strongest on hit generation, while ChemLlama is strongest on specificity. Lead scores differ less across methods than the hit and specificity scores do. The same per-task layout under all-but-me selection is in Appendix~\ref{app:hparam-ablation}.

\subsection{Test Tasks}\label{sec:leaderboard}

\begin{figure*}[t]
\centering
\includegraphics[width=\linewidth]{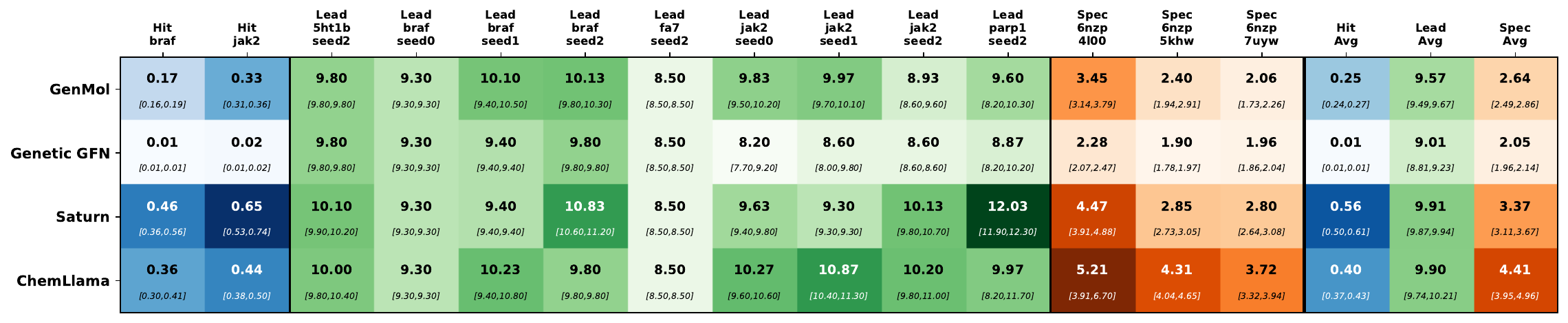}
\caption{Test task leaderboard. Rows: methods, columns: test tasks by category. Hyperparameters selected per Section~\ref{sec:validation-tasks}. Best per task highlighted. Colors are normalized separately by category, so intensities should only be compared within category. More intense colors indicate higher performance relative to other entries in the same category.}
\label{fig:leaderboard}
\end{figure*}

Figure \ref{fig:leaderboard} presents the final evaluation of all the methods under the category-specific rule in Section~\ref{sec:validation-tasks}. Cell annotations report mean scores with $95\%$ bootstrap confidence intervals over evaluation seeds (Appendix~\ref{app:leaderboard-stats}). No method is best on every task. Saturn is considerably better at hit generation, and ChemLlama is better at specificity. On lead optimization the four methods are comparable. On this benchmark, hit finding and selectivity therefore favor different algorithms.




\subsection{Evaluation with Extended Budget}
\label{sec:extended-budget}

\begin{figure}[t]
\centering
\includegraphics[width=\linewidth]{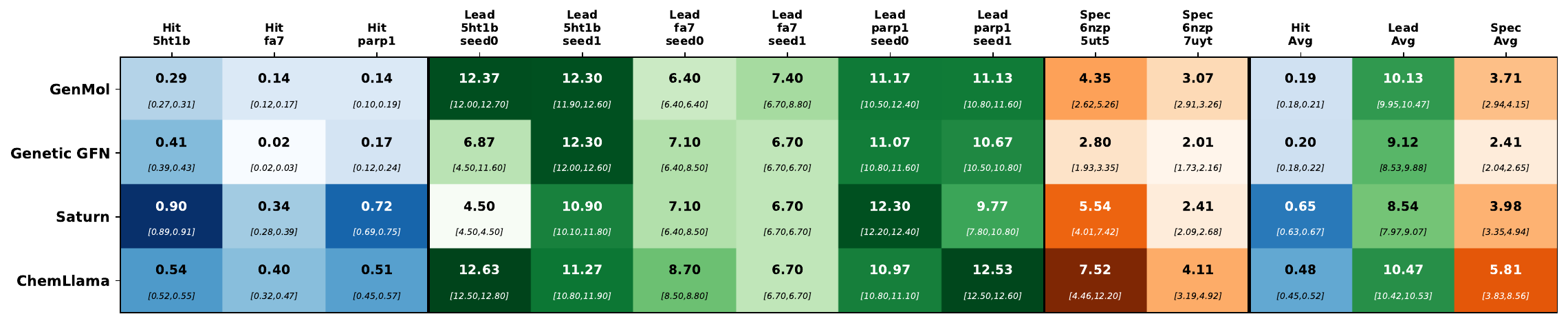}
\caption{Validation performance under per-task best hyperparameters (one row per method). For each task column, the reported value uses the hyperparameter setting that maximizes the validation metric on that task, corresponding to an unrestricted oracle-call scenario for model selection on the validation split. Layout and per-category coloring match Figure~\ref{fig:perf_table}.}
\label{fig:task_best_val}
\end{figure}

In previous sections, we evaluated and analysed the optimization power and generalization of the different methods, under a strict evaluation budget, which emulates a realistic drug design environment. However, it is not uncommon that in practice one is interested in finding a good solution and is less worried about the strict oracle budget. To accomodate these needs, we additionally evaluate each algorithm under an \emph{extended oracle-call} regime: for every task, we select hyperparameters that maximize performance on that same task. This isolates per-task ceiling behavior when model selection is not constrained by cross-task generalization. The results are reported in Figure~\ref{fig:task_best_val}.

\subsection{Statistical Significance}
\label{sec:stats-analysis}

The intervals in Figure~\ref{fig:leaderboard} describe seed-level uncertainty for one method on one task. They do not test whether one method ranks above another across a category. Some of those intervals do not overlap. On hit generation, Saturn and ChemLlama are separated from GenMol and Genetic-guided GFN. On JAK2, Saturn is also separated from ChemLlama, while on BRAF their intervals overlap. On specificity, ChemLlama is separated from GenMol and Genetic-guided GFN on every task. It is also separated from Saturn on 5KHW and 7UYW, while on 4L00 their intervals overlap. ChemLlama's category-average interval is separated from all three methods. On lead optimization, three evaluation seeds and close scores leave most pairwise intervals overlapping. A category-level claim needs a comparison that resamples tasks and seeds together.

\begin{figure}[!htbp]
  \centering
  \includegraphics[width=0.75\linewidth]{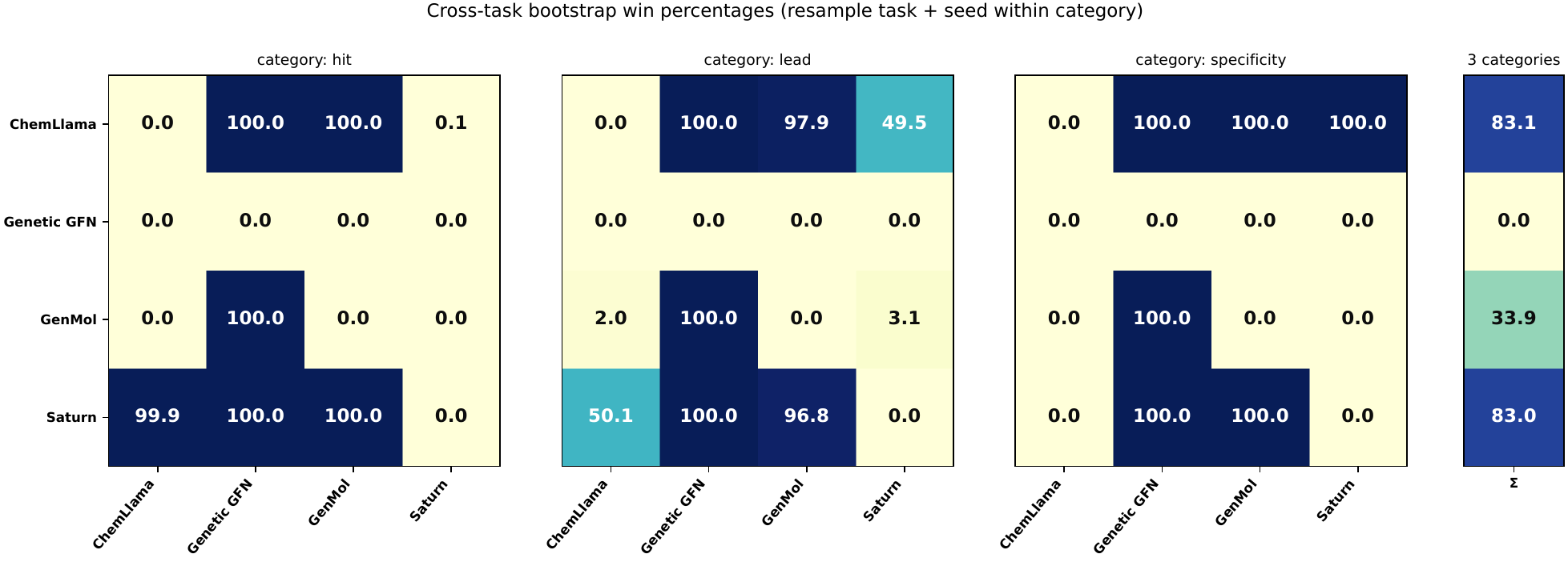}
  \caption{%
    \textbf{Category-level cross-task bootstrap win percentages.}
    \emph{Left three panels:} $4 \times 4$ matrices for hit ($K{=}20$), lead ($K{=}27$), and specificity ($K{=}9$) (task--seed pool sizes).
    Rows and columns are methods. Cell $(i,j)$ is the percentage of $B=10{,}000$ bootstrap replicates in which method~$i$'s resampled mean strictly exceeds method~$j$'s after jointly resampling task blocks and seeds within the category.
    Colour encodes percentage on $[0,100]$.
    \emph{Rightmost panel ($\Sigma$):} per-method mean of the three category-level mean win rates.}
  \label{fig:bootstrap-win-cross-task}
\end{figure}

Figure~\ref{fig:bootstrap-win-cross-task} reports that comparison. For each evaluation category (hit, lead, specificity) we bootstrap over pools of (task, seed) pairs, average each method's score on each draw, and report how often one method's category-level mean strictly exceeds another's across $B=10{,}000$ replicates. This quantifies head-to-head outcomes under task-and-seed resampling within a category. How those confidence intervals are computed, the per-task bootstrap win matrices, and the full resampling procedures are in Appendix~\ref{app:additional-stats}.

\section{Hyperparameter Sensitivity and Transferability}
\label{sec:analysis}

A central question in the evaluation of molecular optimization methods is whether hyperparameters tuned on one set of tasks generalize to new tasks. Robust algorithms should exhibit hyperparameters that work reasonably well on different protein targets and task types. Beyond leaderboard rankings, we study how each method responds to hyperparameter choice and what that implies for deployment on new targets. The PMO-Dock protocol permits extensive hyperparameter sweeps on validation tasks but restricts oracle use on held-out test tasks, since molecular optimization is oracle-limited and such repeated tuning is rarely affordable at scale. We therefore characterize method behavior on the validation split, where tuning is allowed, and ask whether hyperparameters selected there can be reused on related tasks without a new sweep, as required on test. This motivates two complementary analyzes. \emph{Hyperparameter transferability} measures whether the best parameter on one validation task or category remain effective on another. We study transferability along two axes: \textbf{within-category transfer} and \textbf{cross-category transfer}. \emph{Hyperparameter sensitivity} measures how peaked each method's hyperparameter landscape is and how much performance varies when moving away from the task-optimal configuration.




\subsection{Within-category transfer.}

\begin{figure}[!t]
\centering
\includegraphics[width=\linewidth]{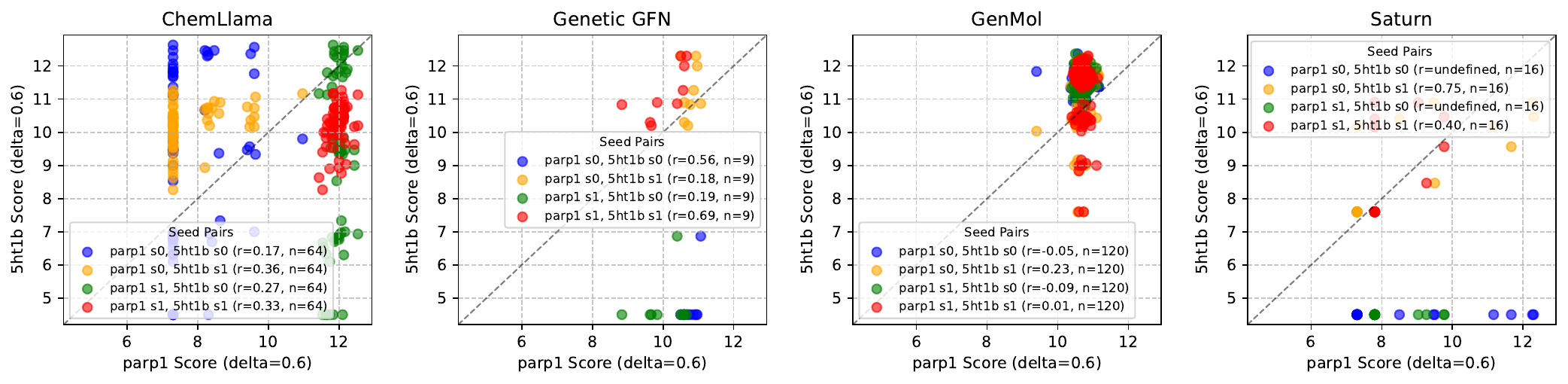}
\caption{Cross-target lead optimization performance (parp1 vs 5ht1b, $\delta=0.6$). One panel per method, colors represent different seed-molecule pairs. Points are aggregated across random seeds, and legends report Pearson correlations and sample sizes. Cross-target lead transfer is generally weak for all four methods.}
\label{fig:lead_cross_target}
\end{figure}

We study whether hyperparameters tuned on one validation target remain effective on another target of the same task type. Strong within-category transfer means one configuration can be reused on other tasks in the same category without a new sweep. We quantify within-category transfer using cross-protein scatter plots on validation tasks (Appendix~\ref{app:transfer}) and the same-category blocks of the hyperparameter transfer matrices (Figures~\ref{fig:transfer_heatmaps} and~\ref{fig:transfer_heatmaps_full}). In the transfer matrix, each cell records the drop in target-task performance when using the source task's best hyperparameters instead of the target's own optimum. In the scatter plots, each point is one swept hyperparameter configuration and the axes compare performance on two targets of the same type. Hit-generation pairs are in Appendix~\ref{app:cross_target}. Lead optimization is shown in Figure~\ref{fig:lead_cross_target}.

Patterns are strongly method-dependent. For hit generation, Saturn shows clear cross-protein alignment, GenMol shows essentially none, and ChemLlama and Genetic-guided GFN are mixed. For lead optimization, transfer is fragile for all methods: hyperparameters selected on one protein rarely predict performance on another (Figure~\ref{fig:lead_cross_target}). In the transfer matrices, Genetic-guided GFN is the least affected by which source task supplied the hyperparameters, while Saturn, ChemLlama, and GenMol degrade more when moving away from the target-specific optimum. Per-pair correlations and full matrix heatmaps are reported in Appendix~\ref{app:transfer} and~\ref{app:cross_target}.

\subsection{Cross-category transfer.}

\begin{figure}[!t]
\centering
\includegraphics[width=\linewidth]{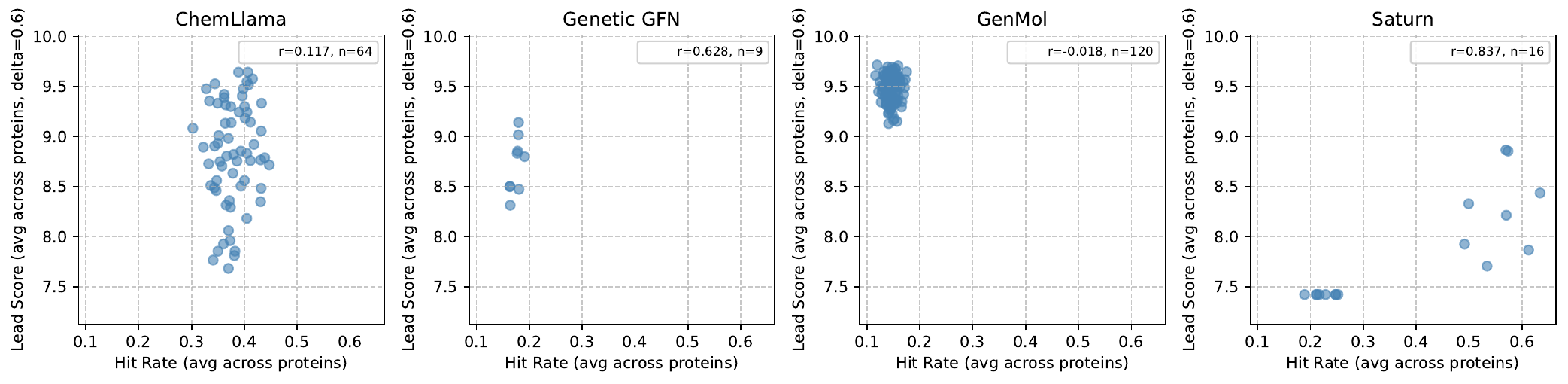}
\caption{Hit vs lead (averaged across proteins, lead at $\delta{=}0.6$). One panel per method, legends report Pearson correlations and sample sizes. Saturn shows the strongest correlation, and GenMol is near zero. Per-protein breakdown in Figure~\ref{fig:hit_vs_lead_protein}.}
\label{fig:hit_vs_lead_avg}
\end{figure}

Cross-category transfer asks whether hyperparameters tuned on one category (hit, lead, or specificity) also perform well on another. This is the question a user faces when tuning on one objective and then running the same settings on a different objective, without a new sweep.
We measure it with a hit-versus-lead scatter plot on the validation tasks (Figure~\ref{fig:hit_vs_lead_avg}; the per-protein breakdown is in Figure~\ref{fig:hit_vs_lead_protein}) and with the off-category blocks of the transfer matrices (Appendix~\ref{app:transfer}). Each point is one hyperparameter configuration. The axes are hit rate and lead score at $\delta=0.6$, either averaged across proteins or shown separately for each protein. A strong positive correlation means configurations that do well in one category also do well in the other.

Saturn has the strongest hit-to-lead correlation, and the correlation is positive on every protein. Genetic-guided GFN is also positively correlated, on a much smaller set of configurations. ChemLlama's correlation is moderate, and GenMol's is near zero. In the transfer matrices, moving across categories costs more than moving within a category for Saturn, ChemLlama, and GenMol. Genetic-guided GFN changes little at the category boundary. We additionally examined specificity-pair transfer and held-out target generalization in Appendices~\ref{app:specificity} and~\ref{app:heldout}.

\subsection{Hyperparameter Sensitivity}

Hyperparameter sensitivity is the drop in performance when a method cannot use the configuration that is best for the task at hand. A sensitive method loses substantial performance under a small change in hyperparameters, or when a different selection rule is used.

We measure this in two ways. First, scatter plots of the swept configurations show whether the same settings do well under two related conditions. Each point is one configuration. A tight, strongly correlated cloud means the preferred configurations stay the same. A dispersed cloud with a weak correlation means they change. Second, we compare validation performance under category-specific selection (Figure~\ref{fig:perf_table}) with validation performance when each task uses its own best hyperparameters (Figure~\ref{fig:task_best_val}). The per-task choice is available only when every task is tuned on its own scores. The gap is the cost of reusing a transferred configuration instead of retuning.

Sensitivity to the lead-optimization constraint is a separate check. Comparing $\delta=0.4$ with $\delta=0.6$ (Appendix~\ref{app:lead_thresh}, Figure~\ref{fig:lead_target_seed}) shows that both the similarity threshold and the hyperparameter choice change the score.

Category-specific selection can also be compared with all-but-me selection (Figure~\ref{fig:hparam_all_but_me}). Saturn gains the most from staying within category. GenMol shows only a small gain, ChemLlama a small loss, and Genetic-guided GFN barely changes, matching its smaller cross-task differences in the transfer matrices. Where transfer is strong and the gap to the per-task best is small, as for Saturn on hit tasks, hyperparameters tuned on similar validation targets can be reused. Where the correlation is weak and the gap is larger, as for GenMol, one configuration should not be reused across categories. Those methods need a validation task of the same category before the test evaluation, and a similar protein when the task is hit generation or lead optimization. Genetic-guided GFN changes little with the source task.

\section{Do these Methods Hack Rewards?}
\label{sec:role-exhaustiveness}
\textbf{The role of docking exhaustiveness.}~Repeated docking of the same protein--ligand pair with QuickVina can yield different scores, which may produce false negatives when only a single run is used. This is especially relevant for reward-hacking risk in specificity tasks. To choose a practical protocol for that stage, we ran a small-scale comparison of replication versus exhaustiveness (multiple independent runs versus raising exhaustiveness). Three independent runs at exhaustiveness $8$ were more stable than one run at exhaustiveness $16$, and agreed closely with a much costlier 3$\times$ex16 scheme, we therefore adopt 3$\times$ex8 for specificity docking. Appendix~\ref{app:exhaustiveness-analysis} provides the study details and Figure~\ref{fig:exhaustiveness-analysis}.
Further checks of the docking oracle, including exhaustiveness $1$ versus $8$ on hit and lead tasks, the agreement of QuickVina with AutoDock Vina, and a PoseBusters assessment of high-scoring poses, are reported in Appendix~\ref{app:additional-docking-oracle}.\\\textbf{Manual inspection.}~To qualitatively assess the generated compounds, we manually inspected molecules produced by the generative models. Across all evaluated methods, the generated structures appeared chemically valid and well-formed. We did not observe unusual artifacts, unrealistic substructures, or chemically implausible patterns in the sampled molecules. Consistent with this observation, the Figure \ref{fig:manual-sensitivity} presents a representative example from the specificity optimization task: a chemically valid molecule generated by ChemLlama. This molecule exhibits the desired binding profile, with favorable binding to the target and weak binding to the anti-target.


\section{Conclusion}

In this paper, we developed a novel benchmark called PMO-Dock, which emulates a real world drug discovery environment and evaluates the ability of molecular optimization algorithms to generalize to previously unseen tasks, under a strict sample efficiency setting. The benchmark consists of 25 problems, composed of hit generation, lead optimization and our newly proposed specificity binding tasks.
We thoroughly evaluated four well-performing optimization algorithms on our benchmark and revealed important insights about their distinct strengths and weaknesses, which ultimately guide their applicability and appropriateness in different drug design workflows. We believe PMO-Dock provides a concrete protocol to support and measure algorithmic advancements in the field.

\section*{Acknowledgements}
This research was supported by the HESC MESCS RA grant No. 24FP-1A058 and NVIDIA Academic Grant Program using NVIDIA A100 80GB GPUs on NVIDIA Brev. 






\bibliographystyle{unsrtnat}
\bibliography{references}

\newpage

\newpage
\appendix
\onecolumn

\section{Limitations}
\label{sec:limitations}

While we believe that our work standardizes the comparison of the docking-based molecular optimization tasks, we recognize two main limitations. First, we used QuickVina throughout the paper for computing binding affinity/docking score, however, we believe that more advanced simulations may be used instead to mitigate any potential hacking of the QuickVina simulation and make the claims and significance of the results more relevant in real-world drug discovery. Secondly, our benchmark mainly focuses on docking-based objectives and problems, however, drug development involves additional important requirements, such as solubility and non-toxicity, which should be incorporated in the future advancements of the benchmark.

\section{Detailed Method Descriptions and Hyperparameter Configurations}
\label{app:methods}

This section provides detailed descriptions of each method evaluated in the benchmark, followed by complete hyperparameter search space specifications.

\subsection{Saturn}
Saturn~\citep{guo2024saturn} is a language-model-based molecular generator trained with reinforcement learning using an autoregressive policy that generates SMILES representations of molecules.
It employs an \emph{Augmented Memory}~\citep{guo2023augmented} replay buffer that stores the top-$K$ high-reward molecules encountered during search (e.g., $K\!=\!100$ in the original work). Buffer entries are repeatedly replayed by training on randomized SMILES augmentations to improve sample efficiency.
A scaffold-based diversity filter (e.g., Bemis–Murcko~\citep{Bemis1996} scaffold tracking) penalizes repeated scaffolds to mitigate mode collapse.

The authors proposed an additional feature, graph-based genetic operators (crossover and mutation) applied to the buffer population to produce new candidate molecules that can refresh the buffer. We kept this genetic algorithm component as a binary hyperparameter. The other hyperparameters we tested are buffer size $\sigma \in \{64, 128, 256, 512\}$ and two reward types: the one borrowed from GEAM algorithm and a hit-based reward. This resulted in 16 hyperparameter combinations.\\\textbf{Reward functions.}~Saturn uses multiplicative reward functions. For \textbf{hit optimization} tasks, the reward is:
\[
R(x) = \frac{s^{(\text{target})}(x)}{20} \cdot \mathrm{QED}(x) \cdot \frac{10 - \mathrm{SA}(x)}{9},
\]
where all components are normalized to $[0, 1]$.

For \textbf{lead optimization} tasks, the reward is:
\[
R(x) = \mathrm{Sim}(x, x_0) \cdot \frac{s^{(\text{target})}(x)}{20} \cdot \mathrm{QED}(x) \cdot \frac{10 - \mathrm{SA}(x)}{9},
\]
where the first term encourages high similarity to the seed molecule $x_0$.

For \textbf{specificity optimization} tasks, Saturn's reward function explicitly penalizes anti-target binding:
\[
R(x) = \left(1 - \frac{s^{(\text{off})}(x)}{20}\right) \cdot \frac{s^{(\text{target})}(x)}{20} \cdot \mathrm{QED}(x) \cdot \frac{10 - \mathrm{SA}(x)}{9},
\]
where the first term $(1 - s^{(\text{off})}(x)/20)$ encourages low anti-target binding.

\subsection{GenMol}
GenMol~\citep{genmol} is a masked discrete-diffusion model operating on a fragment-level representation SAFE~\citep{noutahi2023gotta}, where molecules are decomposed into fragment tokens (e.g., BRICS~\citep{Degen2008} fragments).
The model uses a bidirectional (BERT-like) denoising network and performs non-autoregressive iterative unmasking: starting from an all-masked fragment string, it repeatedly predicts token logits, locks in confident tokens, and repeats until decoding completes.
For optimization, GenMol masks whole fragments and resamples them with the diffusion model. The same model can generate a molecule from scratch or from a partially specified set of fragments, which covers hit generation and lead optimization.

We performed an extensive hyperparameter sweep of 120 combinations for GenMol. We tested masking parameter $\gamma \in \{0, 0.1, 0.2, 0.3, 0.4, 0.5\}$, randomness $r \in \{1, 1.2, 2, 3\}$, and softmax temperature $\tau \in \{0.5, 0.8, 1, 1.2, 1.5\}$.\\\textbf{Reward functions.}~GenMol uses a binary reward structure.
For \textbf{hit} and \textbf{lead optimization} tasks, if molecule $x$ satisfies all hit constraints (as defined in Section~\ref{sec:benchmark}), the reward is the docking score:
\[
R(x) = \begin{cases}
s^{(\text{target})}(x) & \text{if } h(x) = 1 \\
0 & \text{otherwise}
\end{cases},
\]

For \textbf{specificity optimization} tasks, the reward is:
\[
R(x) = \begin{cases}
(s^{(\text{target})}(x) - s^{(\text{off})}(x)) & \text{if } h(x) = 1 \\
0 & \text{otherwise}
\end{cases},
\]
where $h(x)$ is the validity indicator function (equal to 1 if all constraints are satisfied, 0 otherwise).

\subsection{Genetic-guided GFN}
Genetic-guided GFN~\citep{kim2024genetic} pretrains a SMILES generative policy (e.g., an RNN) on large chemical corpora to serve as a prior.
During optimization, it fine-tunes the policy with the GFlowNet~\citep{bengio2021flow} trajectory-balance objective while integrating genetic operators: parent selection, Graph-GA~\citep{Jensen2019} crossover and mutation to produce offspring, and replay-buffered off-policy TB updates.
Training uses rank-weighted sampling from the buffer and a combined loss consisting of the TB-loss plus a KL penalty to the pretrained prior: $\mathcal{L} = \text{TB-loss} + \lambda\,\mathrm{KL}(\pi_{\text{pre}} \,\|\, \pi_{\text{new}})$.
The genetic operators propose high-reward candidates, and trajectory-balance training learns to sample those candidates while retaining some diversity.\\\textbf{Reward functions.}~Genetic-guided GFN is introduced mainly for a single scalar objective. The original paper also uses it for several objectives by combining them into one scalar score with fixed coefficients. We do the same. We use scalar scores $s(x)$ aligned with the PMO-Dock task objectives (Section~\ref{sec:benchmark}) and set $R(x) = \exp(-\beta s(x))$ as in the GFlowNet literature. $\beta$ is an inverse temperature parameter.

The range for $\beta$ is $\{10, 25, 50, 70, 100\}$. Motivated by the observation in the original Genetic-guided GFN paper that reward scaling in multi-objective settings affects the effective inverse temperature, we performed $\beta$ selection using one representative task per task category.
Within each category, the number of objectives is fixed, so the selected $\beta$ was transferred to all tasks in that category.
We found that $\beta=50$ performed best in all categories and fixed it for all tasks.

In addition to $\beta$, we tuned the rank-based replay coefficient and the KL regularization coefficient on validation tasks.
The rank coefficient was selected from $\{0.01, 0.03, 0.05\}$ and the KL coefficient from $\{0.0005, 0.001, 0.005\}$, following the official Genetic-guided GFN hyperparameter search space.
All remaining components and hyperparameters were used without modification.

\subsection{ChemLlama}
ChemLlama is a modification of Chemlactica~\citep{guevorguian2024chemlactica}. Chemlactica is a molecular language model trained on PubChem molecules with the Galactica architecture and an evolutionary algorithm on top of that model. ChemLlama keeps this design, replaces the Galactica backbone with Llama, and uses the language model as a neural mutation and proposal operator inside a population-based optimization loop.
A pool of top-performing molecules is maintained, the LLM is prompted to generate analogs of pool members (e.g., similarity prompts), generated candidates are scored by the oracle, and the pool is updated by retaining top unique molecules.
When improvement stalls, the model is fine-tuned on the pool with prompts that pair each molecule with its score, so that it learns the relation between structure and score. Generation then resumes with the adapted model.
The procedure cycles: LLM-based proposal $\to$ oracle evaluation $\to$ pool update $\to$ occasional fine-tuning.

We use the ChemLlama model and tokenizer at \url{https://huggingface.co/yerevann/ChemLlama-1B-40k-steps}, a checkpoint trained from Llama 3-1B. We tested the following range of hyperparameters: pool size $P \in \{10, 20, 40, 80\}$, number of prompts $I \in \{25, 50, 100, 200\}$, and number of similar molecules in the prompt $S \in \{1, 2, 3, 4\}$.\\\textbf{Reward functions.}~ChemLlama scores a property value $m$ against a target interval $[\ell, u]$. The score is $1$ inside the interval and decays with the distance to the nearest endpoint outside it:
\[
g(m;[\ell,u],\sigma)=
\begin{cases}
1 & \text{if }\ell\le m\le u,\\
\exp\!\left(-\dfrac{1}{2}\left(\dfrac{\min(|m-\ell|,|m-u|)}{\sigma}\right)^{2}\right)
& \text{otherwise.}
\end{cases}
\]
The widths are $\sigma=2$ for the docking score, $\sigma=0.1$ for $\mathrm{QED}$, $\sigma=0.7$ for $\mathrm{SA}$, and $\sigma=0.1$ for similarity. In the rewards below, $x$ denotes a candidate molecule, and each $g$ uses the interval and width of the property it is applied to.

For \textbf{hit optimization} tasks, the reward is
\[
R(x)=g\!\left(s^{(\text{target})}(x)\right)\,g\!\left(\mathrm{QED}(x)\right)\,g\!\left(\mathrm{SA}(x)\right).
\]

For \textbf{lead optimization} tasks, the reward is
\[
R(x)=\frac{s^{(\text{target})}(x)}{20}\,g\!\left(\mathrm{Sim}(x,x_0)\right)\,g\!\left(\mathrm{QED}(x)\right)\,g\!\left(\mathrm{SA}(x)\right),
\]
where $\mathrm{Sim}(x,x_0)$ is the similarity to the seed molecule $x_0$.

For \textbf{specificity optimization} tasks, the reward penalizes anti-target binding through the clipped score gap
\[
\mathrm{gap}(x)=\mathrm{clip}\!\left(\frac{s^{(\text{target})}(x)}{20}-\frac{s^{(\text{off})}(x)}{20},\,0,\,1\right),
\]
\[
R(x)=\mathrm{gap}(x)\,g\!\left(s^{(\text{target})}(x)\right)\,g\!\left(s^{(\text{off})}(x)\right)\,g\!\left(\mathrm{QED}(x)\right)\,g\!\left(\mathrm{SA}(x)\right).
\]

\section{Additional Hyperparameter Transfer Analyses}
\label{app:transfer}

We performed independent hyperparameter sweeps for each validation task and algorithm. For each task $i$, we identified the hyperparameter configuration maximizing performance on that task, then measured its performance on all other tasks $j$. Each matrix entry reports the change in task-$j$ performance when using task-$i$ hyperparameters versus task-$j$-optimal settings (after per-task normalization), so off-diagonal structure summarizes cross-task transfer rather than raw scores. This protocol quantifies whether tuning signals learned on one objective or target are predictive for others and motivates model selection strategies that avoid test-set leakage.

\begin{figure}[!htbp]
\centering
\includegraphics[width=0.49\linewidth]{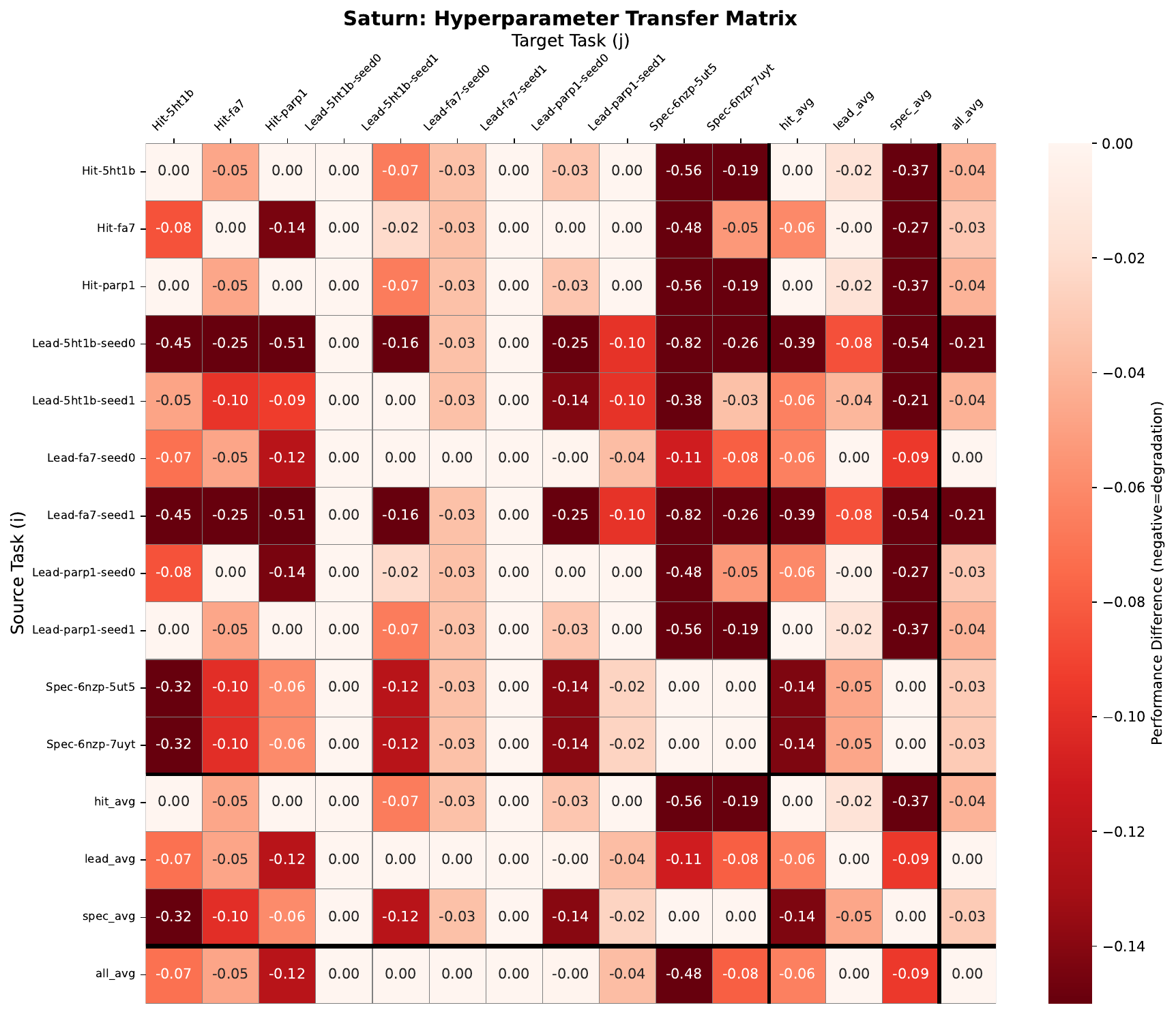}
\includegraphics[width=0.49\linewidth]{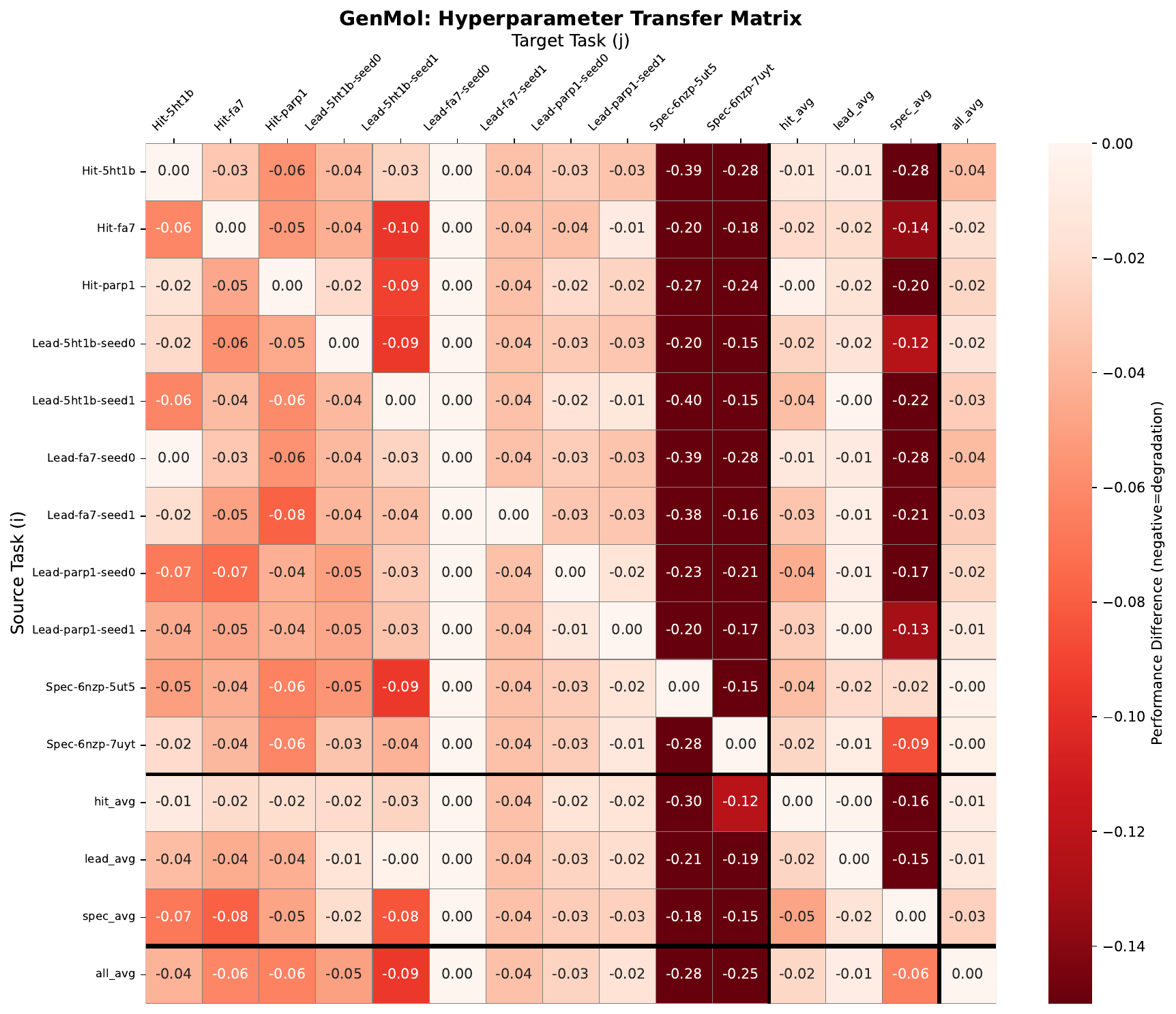}
\caption{Hyperparameter transfer matrices for Saturn (left) and GenMol (right). Cell $(i,j)$ shows the change in task-$j$ performance when using hyperparameters selected from source task $i$ instead of task-$j$-optimal hyperparameters. Scores are normalized by task type before comparison (hit rates direct, lead scores divided by 20, specificity scores divided by 5). Because entries are measured relative to the target-task optimum, values are expected to be non-positive, values near 0 indicate little transfer loss, while more negative values indicate larger degradation. These matrices summarize relative transfer rather than absolute task performance. Thick separators distinguish individual tasks from aggregate rows and columns (hit\_avg, lead\_avg, spec\_avg, all\_avg).}
\label{fig:transfer_heatmaps}
\end{figure}

\begin{figure}[!htbp]
\centering
\includegraphics[width=0.49\linewidth]{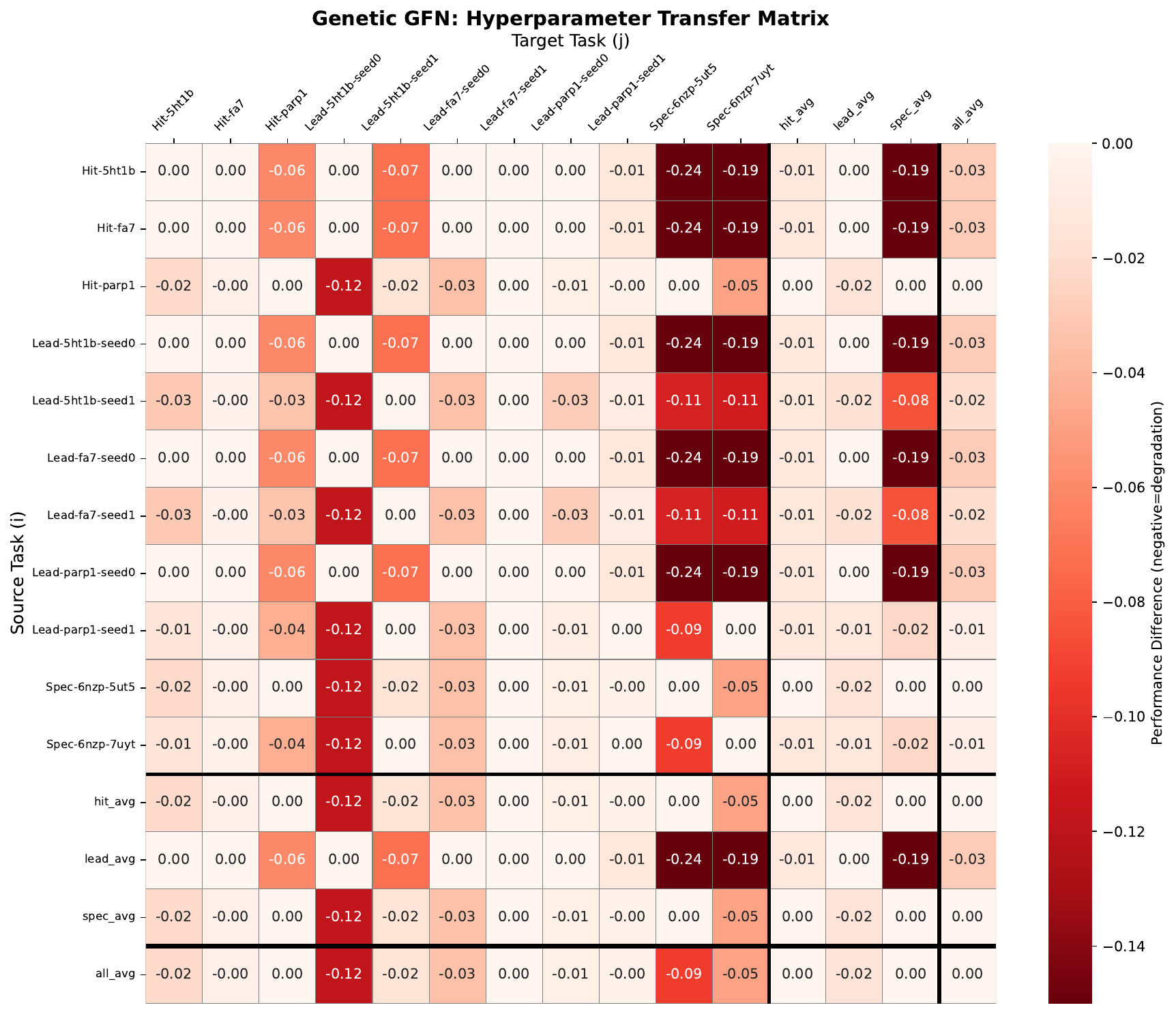}
\includegraphics[width=0.49\linewidth]{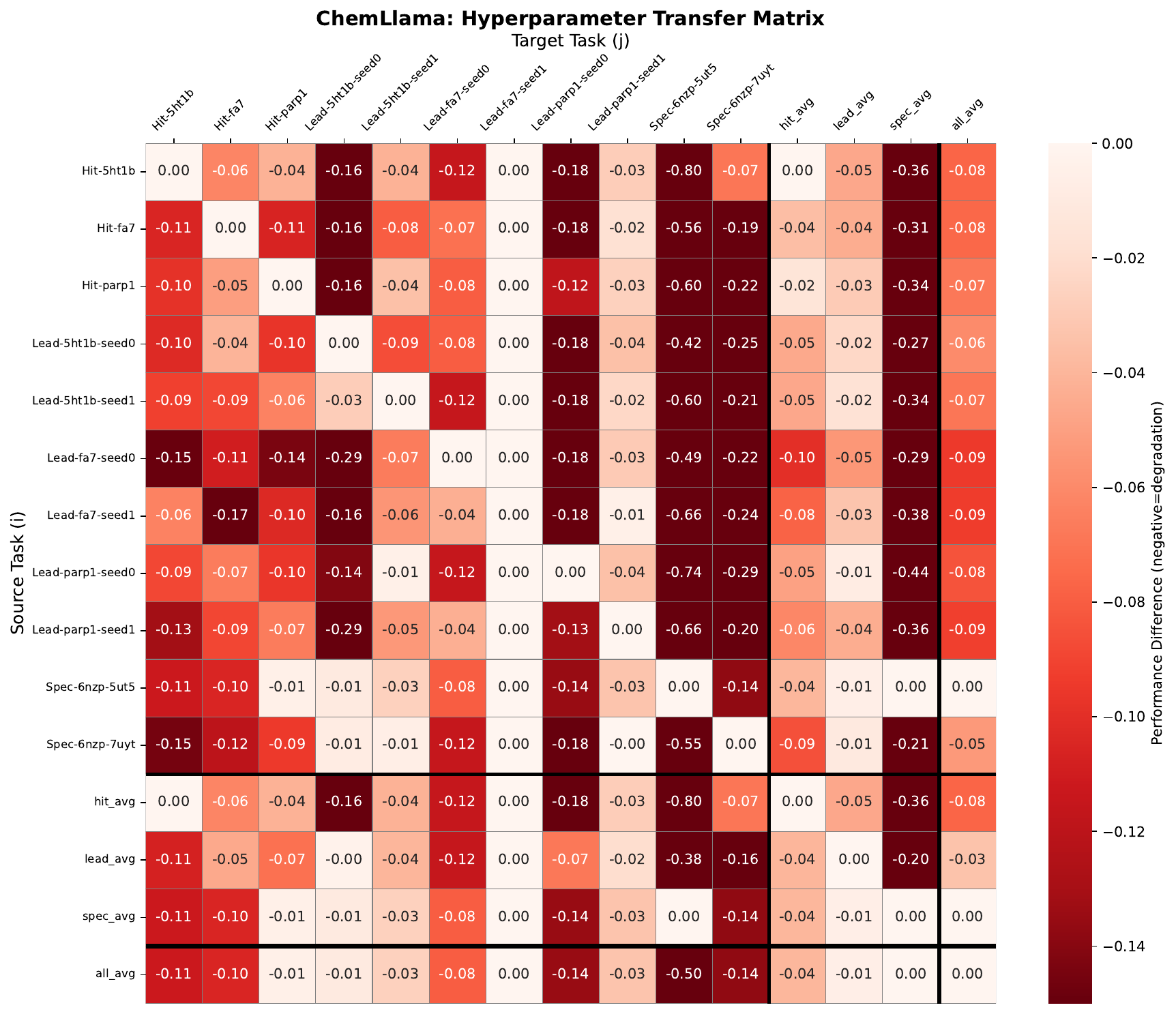}
\caption{Hyperparameter transfer matrices for Genetic-guided GFN (left) and ChemLlama (right). As in Figure~\ref{fig:transfer_heatmaps}, each cell $(i,j)$ reports the change in task-$j$ performance when using hyperparameters selected from source task $i$ instead of task-$j$-optimal hyperparameters. Scores are normalized by task type before comparison (hit rates direct, lead scores divided by 20, specificity scores divided by 5). Values are expected to be non-positive because the reference is the target-task optimum, entries near 0 indicate little transfer loss and more negative entries indicate larger degradation.}
\label{fig:transfer_heatmaps_full}
\end{figure}

Figures~\ref{fig:transfer_heatmaps} and~\ref{fig:transfer_heatmaps_full} show transfer matrices for Saturn, GenMol, Genetic-guided GFN and ChemLlama. Each cell $(i,j)$ reports the relative change in task-$j$ performance when using hyperparameters selected from source task $i$ instead of task-$j$-optimal hyperparameters, after task-type normalization. Thus the heatmaps summarize transfer loss relative to the target-task optimum rather than absolute task performance.

Off the diagonal, Saturn, ChemLlama, and GenMol lose more when the source and target are in different categories than when they share a category. Genetic-guided GFN shows only a small difference between those two cases.

\section{All-but-me hyperparameter selection}
\label{app:hparam-ablation}

\begin{figure}[h]
\centering
\includegraphics[width=\linewidth]{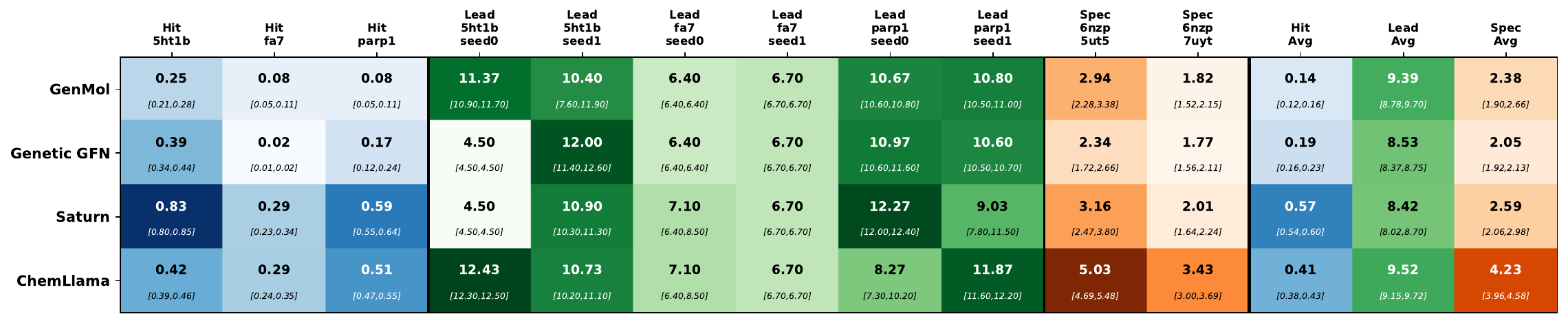}
\caption{Validation performance under all-but-me hyperparameter selection. For each task, hyperparameters maximize average performance on every other validation task. The last three columns are means over the tasks in each category (Hit Avg, Lead Avg, Spec Avg). Colors are normalized separately by category.}
\label{fig:hparam_all_but_me}
\end{figure}

Figure~\ref{fig:perf_table} uses category-specific selection. Figure~\ref{fig:hparam_all_but_me} is the same per-task comparison when each task instead takes the hyperparameters that maximize average performance on all other validation tasks, including tasks from other categories. Layout and coloring match Figure~\ref{fig:perf_table}.

\section{Additional Statistical Analyses}
\label{app:additional-stats}

This appendix reports bootstrap confidence intervals for leaderboard tables and pairwise method comparisons on test tasks. Table cells and aggregate columns in Figures~\ref{fig:perf_table},~\ref{fig:leaderboard}, and~\ref{fig:task_best_val} use the interval definitions in Appendix~\ref{app:leaderboard-stats}. Three further bootstrap analyses on test scores follow: \emph{category-level win matrices} (Figure~\ref{fig:bootstrap-win-cross-task}, summarized in Section~\ref{sec:stats-analysis}) resample (task, seed) pairs jointly within each category, \emph{per-task win matrices} (Figure~\ref{fig:bootstrap-win-per-task}) fix one test task and resample seeds only and \emph{hit-rate distributions} (Figures~\ref{fig:hit-boot-mean} and~\ref{fig:hit-boot-diff}) show the resampled hit rates for the test hit tasks, using the same seed draws as the hit panels in Figure~\ref{fig:bootstrap-win-per-task}.

\subsection{Leaderboard table confidence intervals}
\label{app:leaderboard-stats}

Figures~\ref{fig:perf_table},~\ref{fig:leaderboard}, and~\ref{fig:task_best_val} annotate each cell with the mean score across evaluation seeds for the selected hyperparameter configuration, followed by an equal-tailed $95\%$ bootstrap confidence interval $[\mathrm{lo}, \mathrm{hi}]$ from $B=10{,}000$ seed resamples. Per-task intervals reflect seed-level variability for the reported configuration only. Evaluation seed set sizes are $|\mathcal{S}|=10$ for hit tasks and $|\mathcal{S}|=3$ for lead and specificity tasks. The main tables omit \textbf{Normalized Avg}. \textbf{Hit Avg}, \textbf{Lead Avg}, and \textbf{Spec Avg} are means over tasks in each category, and \textbf{Normalized Avg} is
\[
  \mathrm{Normalized\_Avg} = \frac{1}{3}\left(\mathrm{Hit\_Avg} + \frac{\mathrm{Lead\_Avg}}{20} + \frac{\mathrm{Spec\_Avg}}{5}\right).
\]
For an aggregate column, each bootstrap replicate resamples seeds and recomputes the category mean. The same resampled seed index is used for every method in that category. The Normalized Avg interval applies the formula above to the three category draws from that replicate.

\subsection{Category-level cross-task bootstrap}
\label{app:category-bootstrap}

Figure~\ref{fig:bootstrap-win-cross-task} reports pairwise win percentages under joint resampling of test tasks and seeds within each evaluation category. For each category we form a finite pool of (task block, seed) pairs from the same non-aggregate test blocks used in Figure~\ref{fig:leaderboard}, with pool sizes $K_{\mathrm{hit}}=20$ (two tasks $\times$ ten seeds), $K_{\mathrm{lead}}=27$ (nine tasks $\times$ three seeds), and $K_{\mathrm{spec}}=9$ (three tasks $\times$ three seeds). Across $B=10{,}000$ replicates we draw $K$ pairs uniformly with replacement from that pool, each pair equally likely on every draw and average each method's score over the $K$ drawn (task, seed) units. Cell $(i,j)$ in a category matrix is the percentage of replicates with $\overline{s}_i^{(b)} > \overline{s}_j^{(b)}$ (diagonal zero, ties uncounted). The rightmost $\Sigma$ panel is the per-method mean of the three category-level mean win rates.

\subsection{Per-task seed bootstrap win matrices}

\begin{figure}[!htbp]
  \centering
\includegraphics[width=\linewidth]{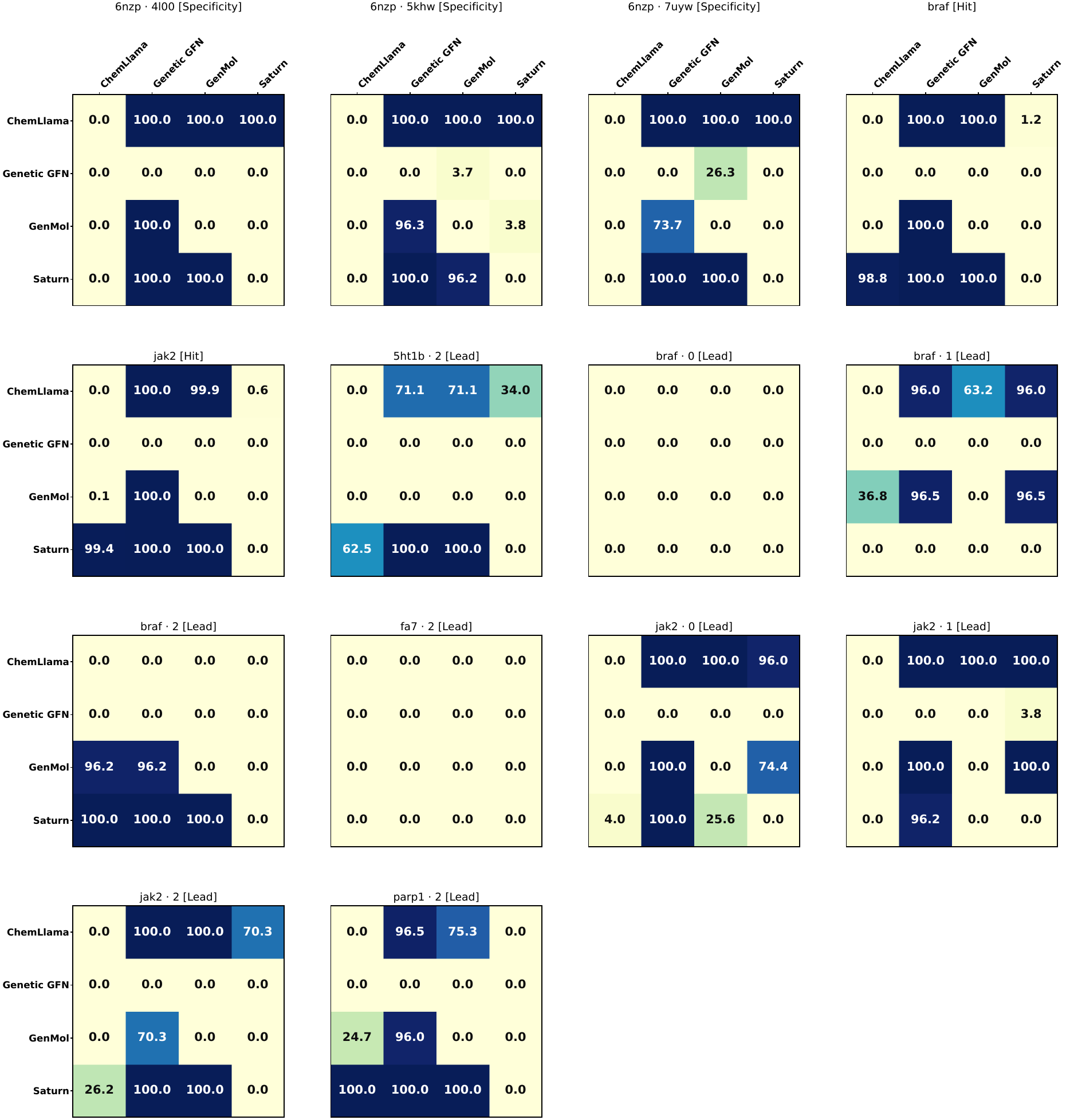}
  \caption{%
    \textbf{Per-task bootstrap win percentages.}
    One $4 \times 4$ heatmap per test task (fourteen panels: two hit, nine lead, three specificity).
    Rows and columns are methods (ChemLlama, Genetic-guided GFN, GenMol, Saturn), cell $(i,j)$ is the percentage of $B = 10{,}000$ bootstrap replicates in which the row method's resampled mean strictly exceeds the column method's under seed resampling for that fixed task.
    Each replicate resamples seeds with replacement from the task's empirical seed set and averages each method's score over the resampled seeds.
    Colour encodes percentage on a common $0$--$100\%$ scale across panels (darker = higher).}
  \label{fig:bootstrap-win-per-task}
\end{figure}

This analysis complements the category-level bootstrap above by fixing one test task and resampling only its random seeds. For every non-aggregate test block (two hit, nine lead, three specificity) and each of $B = 10{,}000$ replicates, we draw $|\mathcal{S}|$ seeds with replacement from that block's empirical seed set, average every method's score over the resampled seeds, and form pairwise win percentages. The resulting fourteen $4 \times 4$ matrices in Figure~\ref{fig:bootstrap-win-per-task} quantify how stable the per-task ranking is to seed variation.

\subsection{Hit-rate bootstrap on test hit tasks}
\label{app:hit-bootstrap}

For test hit tasks we additionally report bootstrap \emph{distributions} of mean hit rate and paired Saturn-minus-competitor contrasts. We also include a \textsc{both} summary that averages BRAF and JAK2 hit rates at each $(m,s)$ before resampling. We use Saturn as the paired reference because it leads on hit tasks in the test leaderboard.

For each task summary $t \in \{\textsc{braf},\,\textsc{jak2},\,\textsc{both}\}$ we collect the observed hit rate $x_{m,s}$ for every method $m$ and evaluation seed $s$ in the empirical seed set $\mathcal{S}$. Across $B = 10{,}000$ replicates we draw $|\mathcal{S}|$ seeds independently with replacement from $\mathcal{S}$ and form
\begin{equation}
  \bar{x}_m^{(b)} = \frac{1}{|\mathcal{S}|}\sum_{k=1}^{|\mathcal{S}|} x_{m,\,s^{(b)}_k},  \qquad
  \bar{\Delta}_{m}^{(b)} = \frac{1}{|\mathcal{S}|}\sum_{k=1}^{|\mathcal{S}|} \bigl(x_{\mathrm{Saturn},\,s^{(b)}_k}-x_{m,\,s^{(b)}_k}\bigr),
\end{equation}
i.e.\ the bootstrap replicate of method $m$'s mean hit rate and of its paired contrast against Saturn on the same resampled seeds. We summarize each bootstrap distribution by its mean and by the equal-tailed $95\%$ percentile interval $\mathrm{CI}_{95}(\cdot)$.

\begin{figure}[!htbp]
  \centering
   \includegraphics[width=\linewidth]{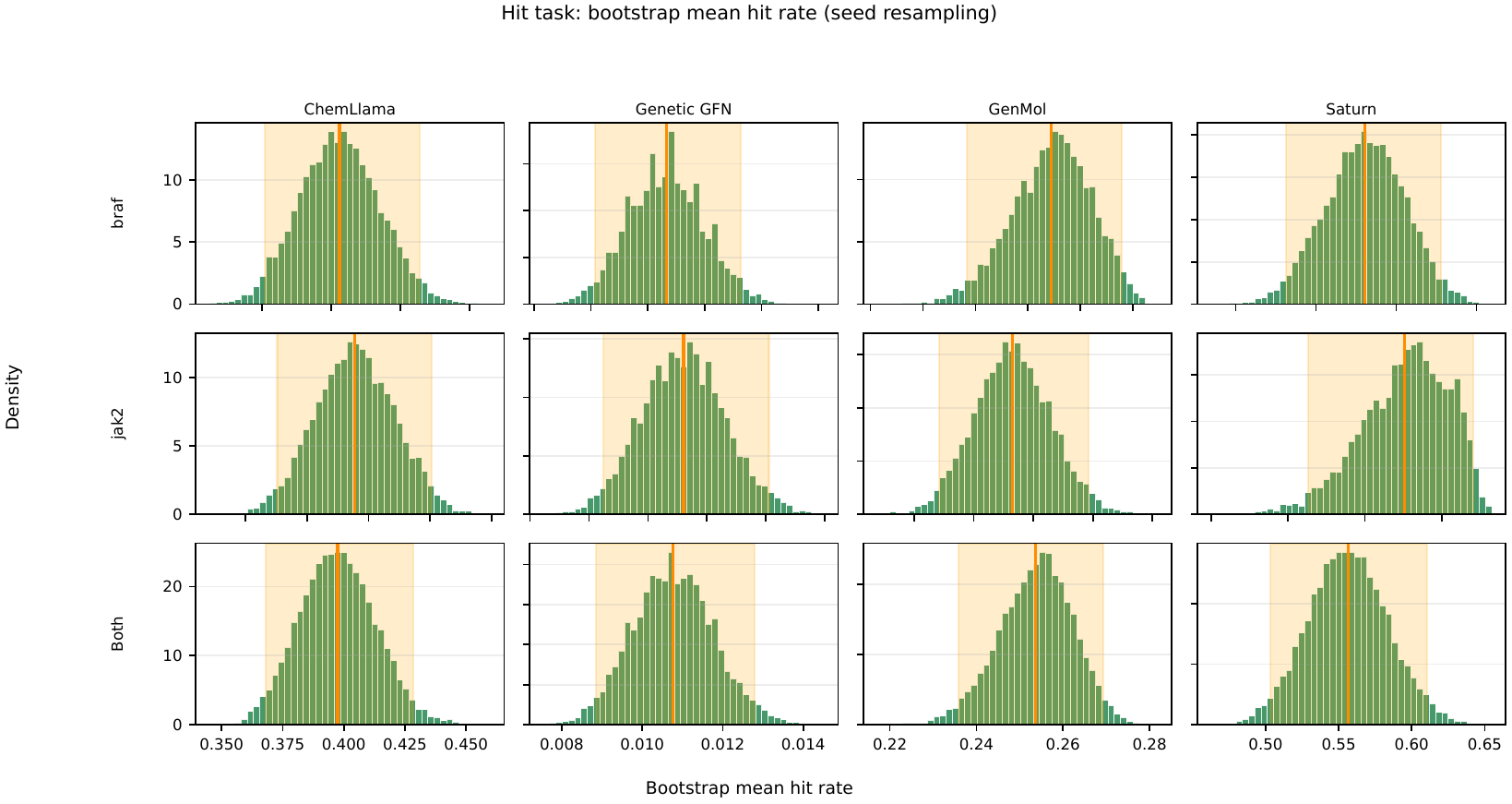}
  \caption{%
    \textbf{Bootstrap distributions of mean hit rate under seed resampling.}
    Rows are the three task summaries (\textsc{braf}, \textsc{jak2}, \textsc{both}), columns are methods. Each panel shows a normalized histogram of $\bar{x}_m^{(b)}$ over $B = 10{,}000$ replicates, the vertical orange line is the bootstrap mean and the shaded band is $\mathrm{CI}_{95}(\bar{x}_m)$.}
  \label{fig:hit-boot-mean}
\end{figure}

Figure~\ref{fig:hit-boot-mean} reports the bootstrap distribution of $\bar{x}_m^{(b)}$ for every (method, task-summary) cell. A wider histogram means the mean hit rate changes more when different seeds are drawn.

\begin{figure}[!htbp]
  \centering
  \includegraphics[width=\linewidth]{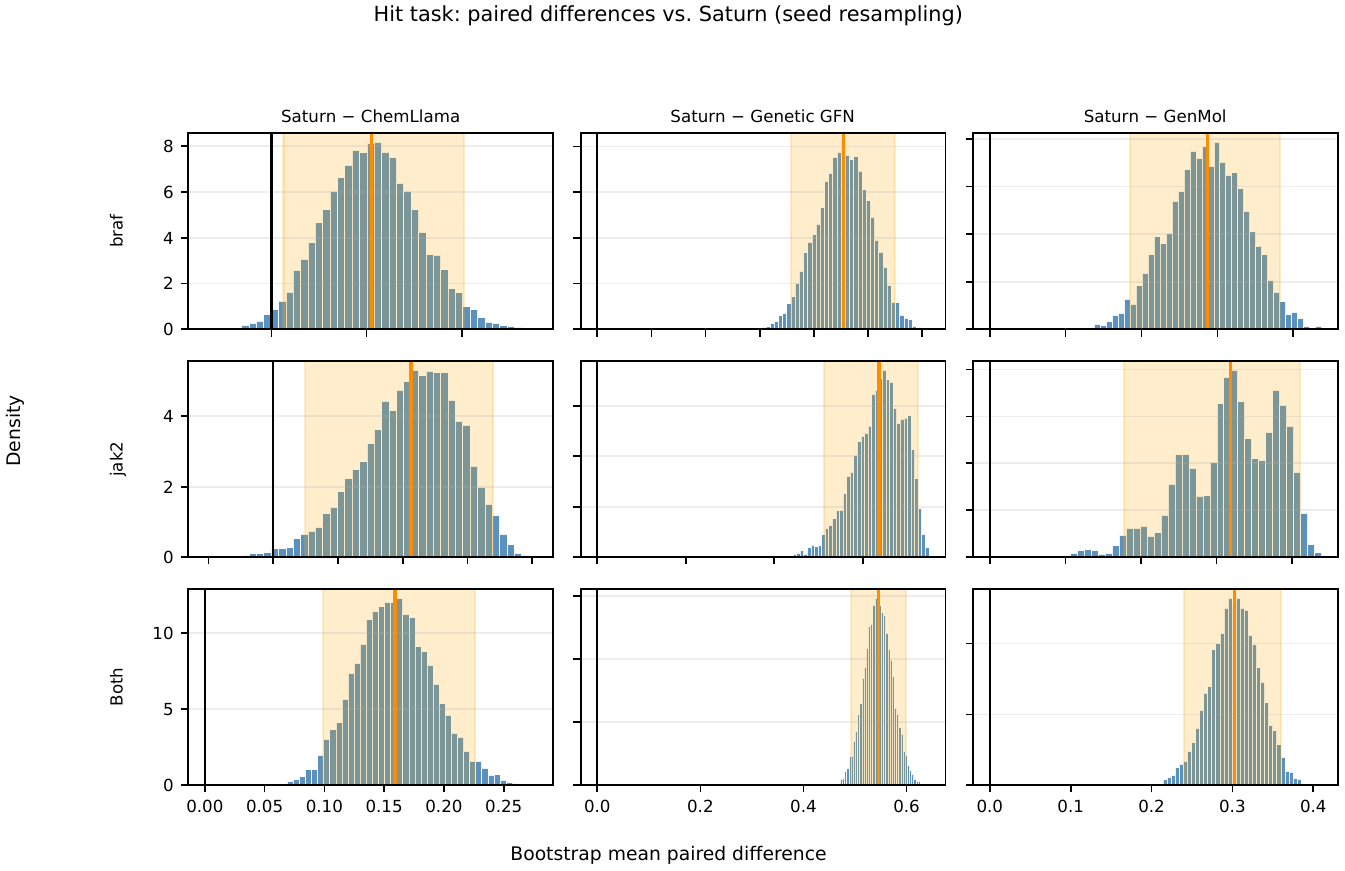}
  \caption{%
    \textbf{Bootstrap distributions of paired mean differences (Saturn $-$ competitor).}
    Rows are the three task summaries (\textsc{braf}, \textsc{jak2}, \textsc{both}), columns are competitors $m \neq \text{Saturn}$. Each panel histograms $\bar{\Delta}_{m}^{(b)}$ over $B = 10{,}000$ replicates. The black line marks zero, the orange line marks the bootstrap mean, and the shaded band is $\mathrm{CI}_{95}(\bar{\Delta}_{m})$.}
  \label{fig:hit-boot-diff}
\end{figure}

Figure~\ref{fig:hit-boot-diff} reports the paired contrast $\bar{\Delta}_{m}^{(b)}$ for every competitor $m \neq \text{Saturn}$ and every task summary. For all reported pairs $(m,t)$ the interval $\mathrm{CI}_{95}(\bar{\Delta}_{m})$ lies strictly above zero, consistent with Saturn winning most bootstrap replicates against each competitor in the hit panels of Figures~\ref{fig:bootstrap-win-cross-task} and~\ref{fig:bootstrap-win-per-task}.

\section{Additional Hyperparameter Transfer Scatter Analyses}
\label{app:transfer-scatter}

This appendix supplements Section~\ref{sec:analysis} and the transfer matrices in Appendix~\ref{app:transfer} with validation-task scatter plots. Each point is one hyperparameter configuration from the validation sweep, axes compare performance on two tasks, targets, or constraint settings. Legends report Pearson correlations and sample sizes.

\subsection{Cross-Target Hit Generation: Additional Protein Pairs}
\label{app:cross_target}

\begin{figure}[!htbp]
\centering
\includegraphics[width=\linewidth]{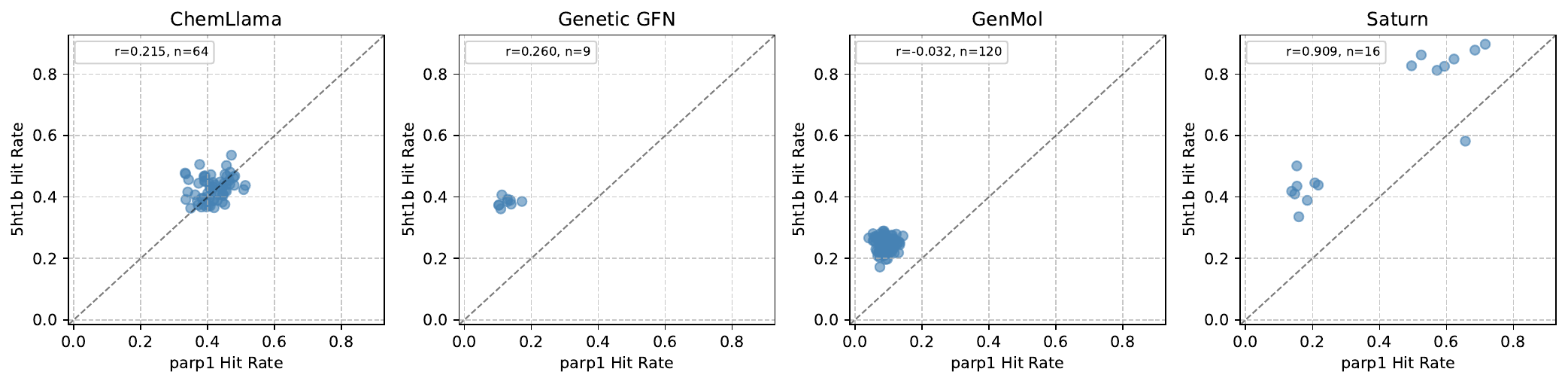}
\caption{Cross-target hit generation performance: parp1 vs 5ht1b. Each point represents a hyperparameter configuration, averaged across random seeds. One panel per method, legends report Pearson correlations and sample sizes.}
\label{fig:hit_cross}
\end{figure}

\begin{figure}[!htbp]
\centering
\includegraphics[width=\linewidth]{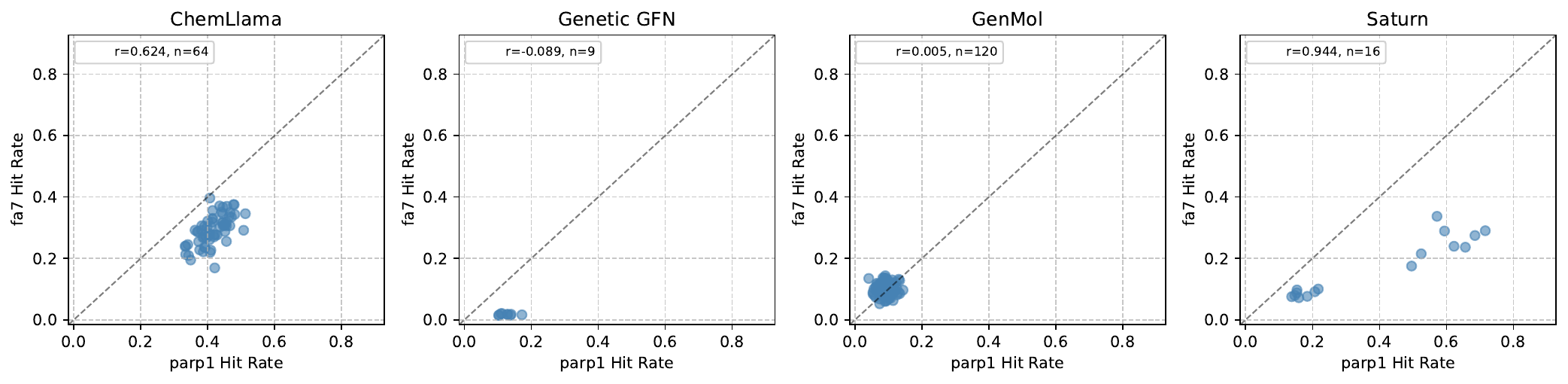}
\caption{Cross-target hit generation performance: parp1 vs fa7. Each point represents a hyperparameter configuration, averaged across random seeds.}
\label{fig:hit_parp1_fa7}
\end{figure}

\begin{figure}[!htbp]
\centering
\includegraphics[width=\linewidth]{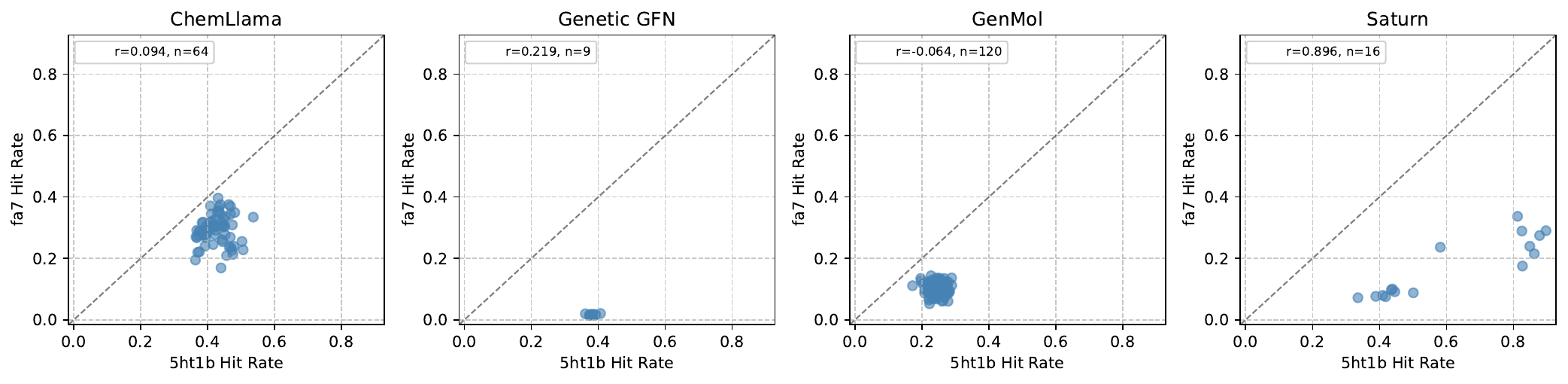}
\caption{Cross-target hit generation performance: 5ht1b vs fa7. Each point represents a hyperparameter configuration, averaged across random seeds.}
\label{fig:hit_5ht1b_fa7}
\end{figure}

Figures~\ref{fig:hit_cross},~\ref{fig:hit_parp1_fa7}, and~\ref{fig:hit_5ht1b_fa7} show hit-generation performance correlations across protein pairs (one subplot per method). Saturn's configurations are strongly correlated on all pairs ($r \in [0.90,\,0.94]$, $n{=}16$). ChemLlama's correlation is moderate on parp1--fa7 ($r{=}0.62$) and weak on 5ht1b--fa7 ($r{=}0.09$). Genetic-guided GFN shows mixed low-$n$ correlations ($n{=}9$). GenMol remains near zero on all pairs ($n{=}120$). These patterns support the within-category transfer discussion in Section~\ref{sec:analysis}.

\FloatBarrier

\subsection{Hit versus lead by protein}
\label{app:hit-lead-protein}

\begin{figure}[!htbp]
\centering
\includegraphics[width=\linewidth]{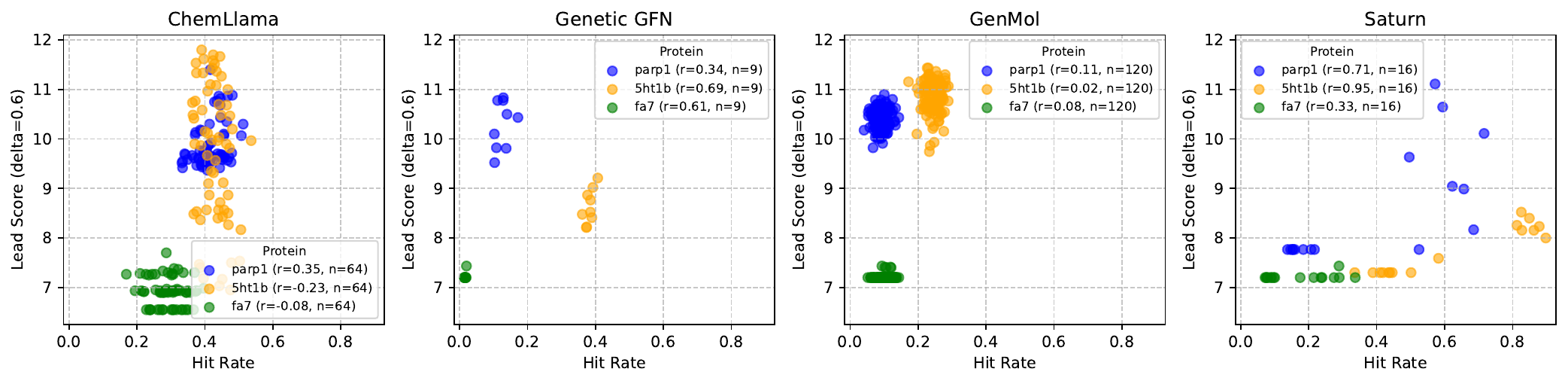}
\caption{Hit rate versus lead optimization score ($\delta=0.6$) by protein. Each protein is shown in a different color within each method panel. Scores are aggregated across random seeds and seed molecules (for lead tasks), legends report Pearson correlations and sample sizes. Saturn's configurations are positively correlated on all three targets. The other methods vary.}
\label{fig:hit_vs_lead_protein}
\end{figure}

Figure~\ref{fig:hit_vs_lead_protein} splits the averaged hit-versus-lead comparison in Figure~\ref{fig:hit_vs_lead_avg} by protein. Saturn's configurations are positively correlated on all three targets. The other methods vary.

\FloatBarrier

\subsection{Lead Optimization: Threshold Transfer}
\label{app:lead_thresh}

\begin{figure}[!htbp]
\centering
\includegraphics[width=\linewidth]{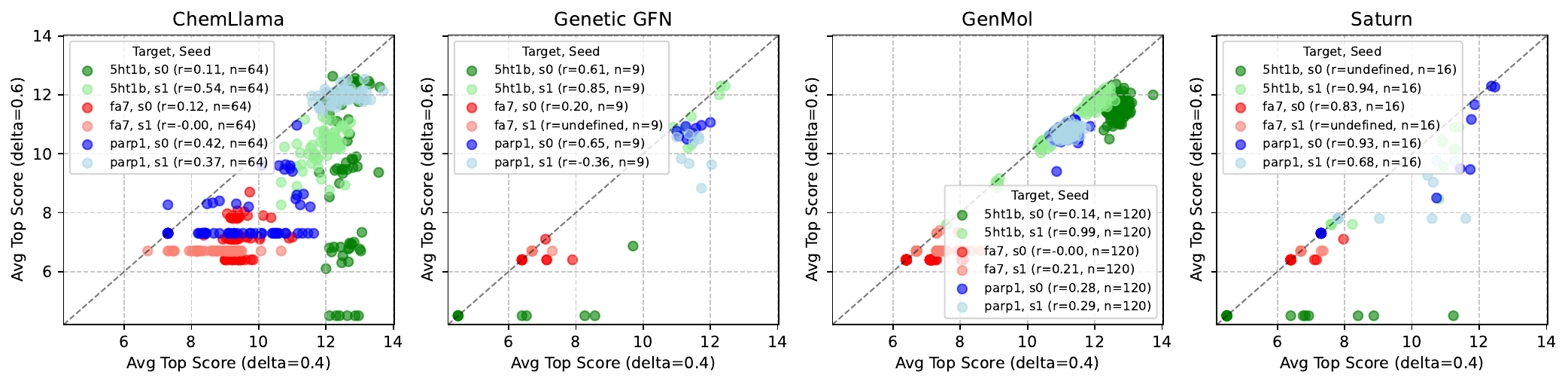}
\caption{Lead optimization performance comparison across threshold values ($\delta=0.4$ vs $\delta=0.6$). Each color represents a specific (target, seed molecule) pair. Legends report Pearson correlations and sample sizes. Labels showing $r=\text{undefined}$ correspond to subsets where one threshold produces constant scores across experiments, making Pearson correlation undefined.}
\label{fig:lead_target_seed}
\end{figure}

Figure~\ref{fig:lead_target_seed} examines transfer when varying the similarity threshold $\delta$ in lead optimization tasks. Each color represents a different (target, seed molecule) combination. The scatter reveals that hyperparameters optimized for $\delta=0.4$ do not reliably transfer to $\delta=0.6$, indicating sensitivity to constraint tightness. In a few subsets the reported Pearson correlation is undefined because one threshold yields constant scores across experiments, so the variance needed for correlation is zero.

\FloatBarrier

\subsection{Specificity Task Transfer}
\label{app:specificity}

\begin{figure}[!htbp]
\centering
\includegraphics[width=\linewidth]{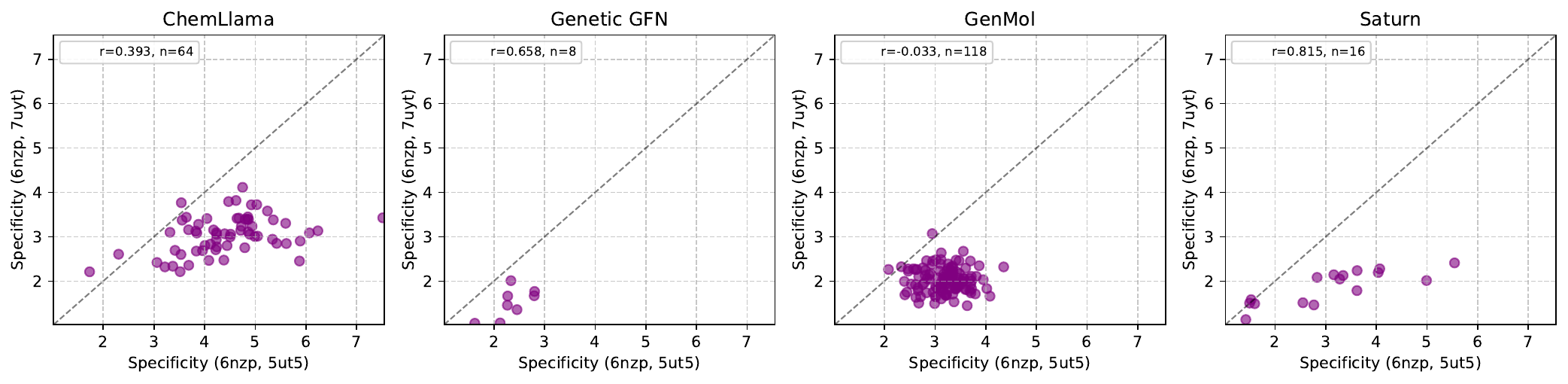}
\caption{Comparison of specificity task performance (6nzp-5ut5 vs 6nzp-7uyt). Axes: mean top-10 specificity score (delta) for each pair. Legends report Pearson correlations and sample sizes. The correlation depends on the method, so a strong score against one off-target does not imply a strong score against the other.}
\label{fig:spec_comparison}
\end{figure}

Figure~\ref{fig:spec_comparison} compares performance on two \emph{validation} specificity tasks (6nzp-5ut5 vs 6nzp-7uyt). The specificity score is computed as $\text{dock}_{\text{target}} - \text{dock}_{\text{off-target}}$. Correlation between the two validation pairs varies by method (more positive for Saturn and Genetic-guided GFN, weaker for ChemLlama, and near zero for GenMol), so strong performance on one pair does not guarantee strong performance on the other.

\FloatBarrier

\subsection{Held-Out Target Generalization}
\label{app:heldout}

\begin{figure}[!htbp]
\centering
\includegraphics[width=\linewidth]{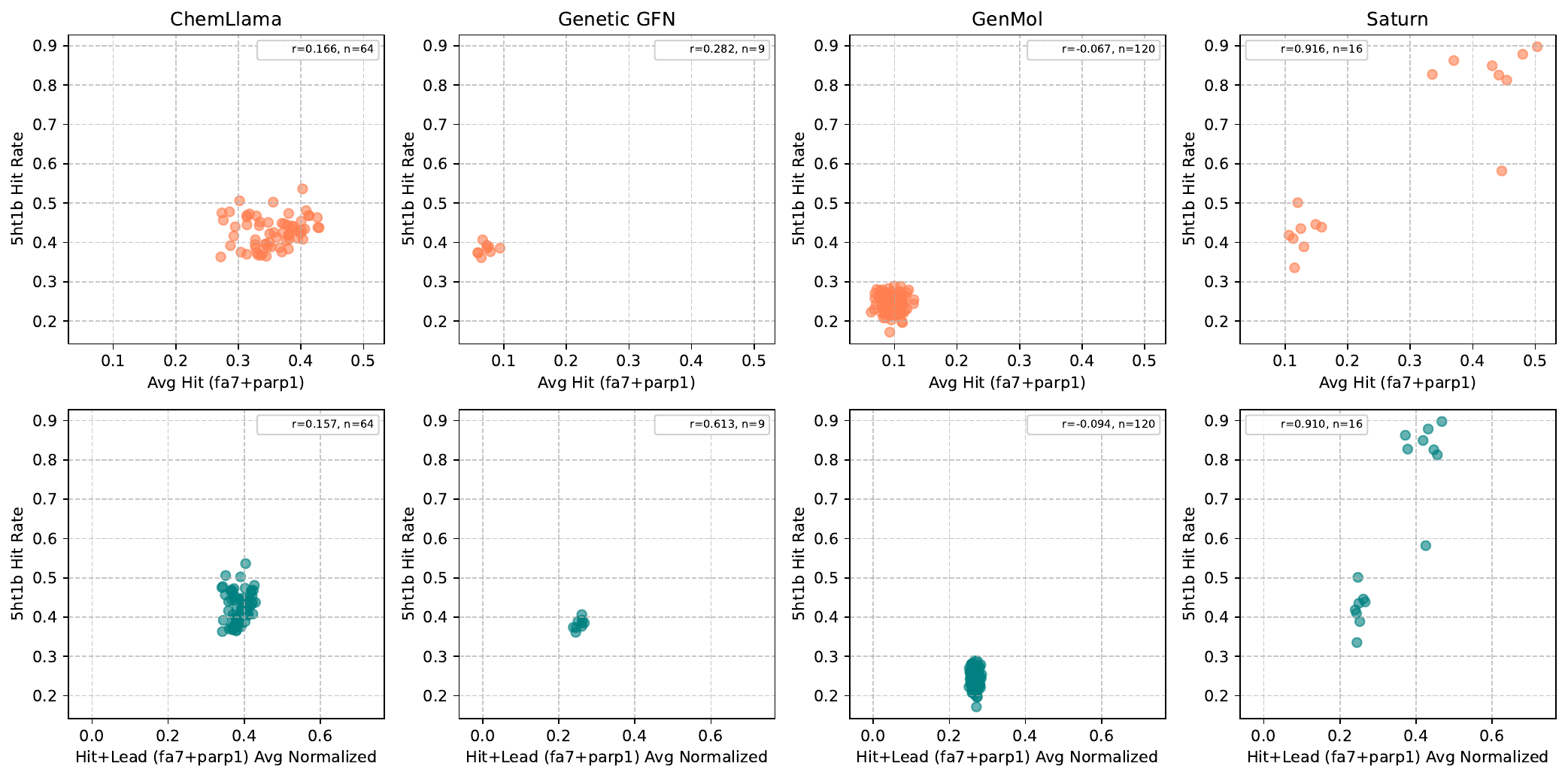}
\caption{Held-out target transfer analysis: 5ht1b hit performance versus other validation tasks. Top row: 5ht1b hit vs average hit on fa7+parp1. Bottom row: 5ht1b hit vs combined hit+lead on fa7+parp1 (normalized). Each column represents one algorithm. Legends report Pearson correlations and sample sizes. Saturn's correlation is substantially stronger than that of the other methods.}
\label{fig:5ht1b_transfer}
\end{figure}

Figure~\ref{fig:5ht1b_transfer} examines whether hyperparameters optimized on fa7 and parp1 transfer to 5ht1b as a held-out target. The 2×4 grid shows 5ht1b hit performance on the Y-axis throughout. In the top row, the X-axis shows average hit performance on fa7 and parp1. In the bottom row, the X-axis shows combined hit and lead performance (normalized) on fa7 and parp1. This simulates the realistic scenario where validation tasks guide hyperparameter selection and test tasks measure true generalization. Saturn's validation aggregates are positively correlated with the 5ht1b hit rate. The other methods are not, so a high aggregate on fa7 and parp1 does not by itself predict the held-out 5ht1b hit rate.

\FloatBarrier

\section{Details of the Docking Pipeline for Specificity Tasks}
\label{sec:docking-details}

Hit and lead tasks use QuickVina with exhaustiveness $1$ as stated in Section~\ref{sec:benchmark}. The specificity tasks use the same scoring function but a separate docking pipeline with higher exhaustiveness for the larger binding boxes described below.

Docking evaluations were performed using QuickVina~\citep{alhossary2015quickvina}. For all five specificity tasks in the benchmark (two validation, three test), the binding box was defined around the ligands observed in the crystallographic structures, ensuring that all amino acid residues participating in ligand interactions were included. The grid box volume did not exceed 27,000 \r{A}$^3$ in any case. Based on this volume, the exhaustiveness parameter was set to 8, according to the Vina manual. 
All remaining preprocessing steps were performed following a standard protocol. Non-essential molecules - including crystallographic water molecules and co-crystallized ligands were removed from the structures. Structures were then analyzed for the presence of alternative atom conformations (altlocs). In cases where alternative positions were observed, the conformation with the highest occupancy was selected. Structures were further checked for missing atoms, and all identified missing atoms were reconstructed (the files are available on GitHub). 
Hydrogen atoms were added using AutoDock Tools~\citep{morris2009autodock}, retaining only polar hydrogens to preserve hydrogen-bonding interactions, and Gasteiger partial charges were subsequently assigned to the protein.

To verify the reliability of our docking pipeline for the current specificity tasks, we tested it on a well-characterized reference compound, Deucravacitinib (BMS-986165) - a first-in-class, approved, highly selective TYK2 pseudokinase inhibitor. The docking results were consistent with experimental data: this molecule exhibited the highest docking score for TYK2 JH2 while showing lower docking scores for TYK2 JH1 as well as JAK2 JH1 and JH2. This validation confirms that our docking setup can reproduce known binding preferences and reliably assess target selectivity. 

\subsection{Exhaustiveness sensitivity (specificity docking)}
\label{app:exhaustiveness-analysis}

This appendix supports the protocol summarized in Section~\ref{sec:role-exhaustiveness} for the \emph{specificity} docking pipeline (Appendix~\ref{sec:docking-details}), where we use exhaustiveness $8$ and aggregate multiple independent runs.

To improve reliability and reduce false negatives, stochastic docking can be stabilized in two ways: increase exhaustiveness, or increase the number of independent runs. In small-scale tests on this pipeline, repeating the calculation with a 3$\times$ex8 scheme produced more stable results than a single run with exhaustiveness $16$. Results from 3$\times$ex8 and 3$\times$ex16 did not differ substantially. Because raising exhaustiveness greatly increases time and compute, specificity docking uses three independent runs at exhaustiveness $8$: more stable than one run at exhaustiveness $16$, and much cheaper than three runs at exhaustiveness $16$.

\begin{figure}[t]
  \centering
  \includegraphics[width=\linewidth]{{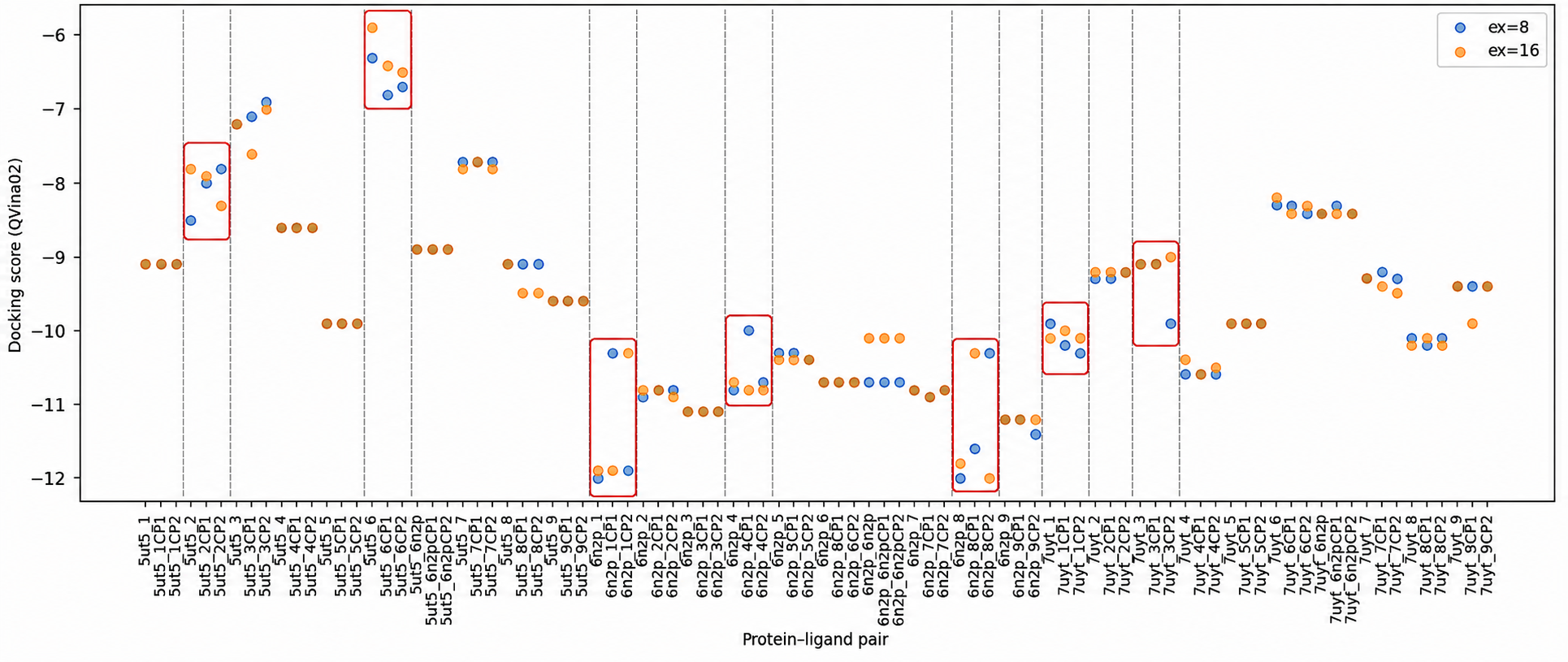}}
  \caption{Exhaustiveness and replication analysis for specificity docking. Red boxes highlight cases where variability across individual runs most strongly affected the aggregated outcome.}
  \label{fig:exhaustiveness-analysis}
\end{figure}

\begin{figure}[t]
  \centering
  \includegraphics[width=\linewidth]{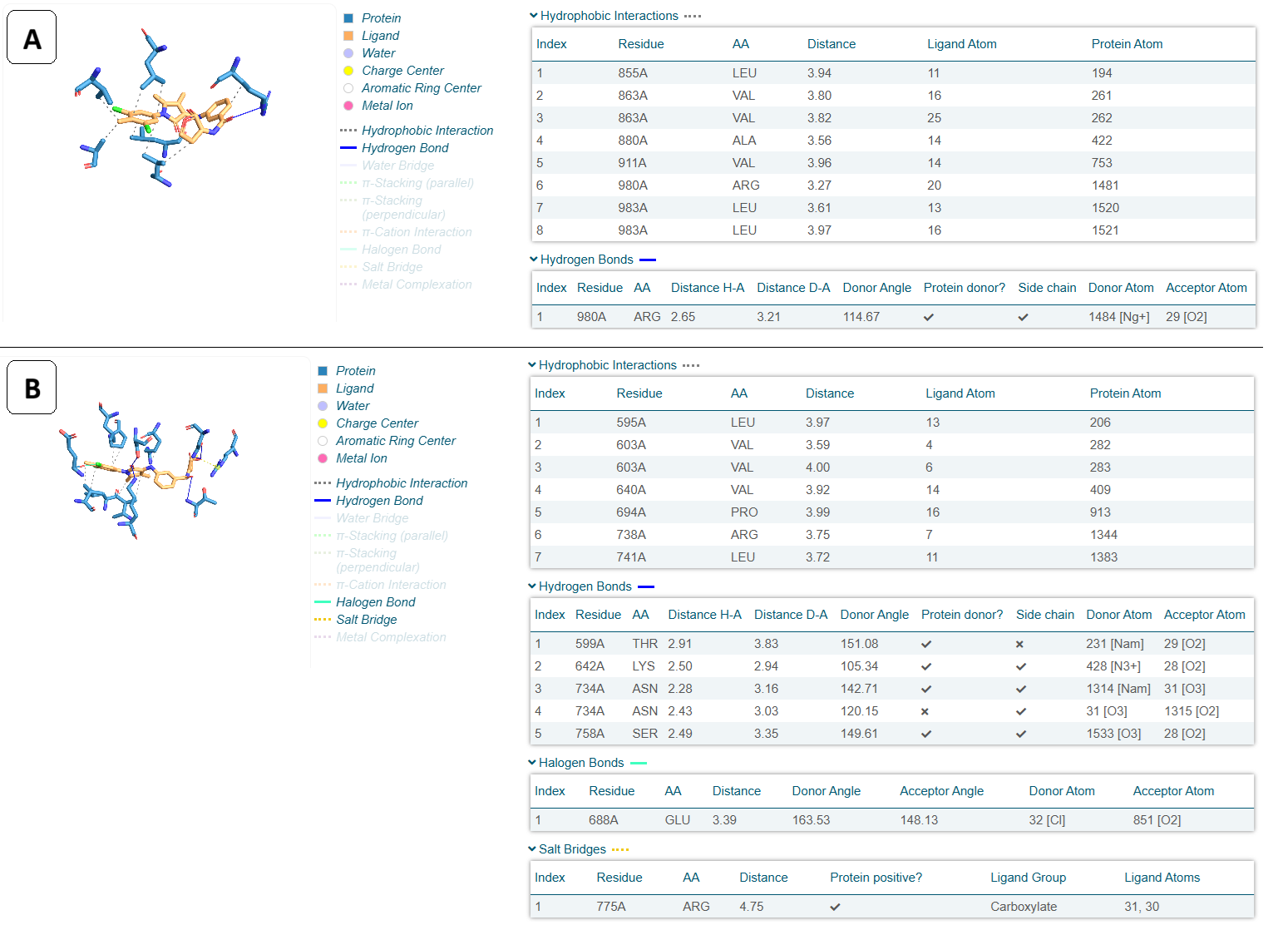}
  \caption{Representative binding poses of a molecule generated by ChemLlama from the specificity optimization task. A) Binding pose of the generated molecule in complex with the anti-target TYK2 JH1 (7UYT). B) Binding pose of the same molecule in complex with the target TYK2 JH2 (6NZP).}
  \label{fig:manual-sensitivity}
\end{figure}


\section{Additional Docking Oracle Analysis}
\label{app:additional-docking-oracle}

This appendix reports additional analyses of the QuickVina docking oracle.

\subsection{Limited scoring-function precision at exhaustiveness 1}
\label{app:exhaustiveness-hit-lead}

We run the optimization with ChemLlama on hit generation (targets: PARP1, FA7, 5HT1B) and lead optimization (targets: PARP1, FA7, 5HT1B) using QuickVina with exhaustiveness $8$, and compare the results with exhaustiveness $1$. Tables~\ref{tab:exh-hit-rate} and~\ref{tab:exh-hit-top1} present the hit-generation results, and Table~\ref{tab:exh-lead-top1} presents the lead-optimization results, across the three target proteins.

\begin{table}[htbp]
\centering
\caption{Hit generation hit rates for ChemLlama at QuickVina exhaustiveness $1$ versus $8$. Entries are mean $\pm$ standard deviation.}
\label{tab:exh-hit-rate}
\begin{tabular}{lcc}
\toprule
Target & Hit rate (ex$=1$) & Hit rate (ex$=8$) \\
\midrule
PARP1 & $0.4570 \pm 0.0791$ & $0.5534 \pm 0.0901$ \\
FA7   & $0.3693 \pm 0.0774$ & $0.3776 \pm 0.0917$ \\
5HT1B & $0.4681 \pm 0.0662$ & $0.6405 \pm 0.0476$ \\
\bottomrule
\end{tabular}
\end{table}

\begin{table}[htbp]
\centering
\caption{Hit generation top-1 docking scores for ChemLlama at QuickVina exhaustiveness $1$ versus $8$. Entries are mean $\pm$ standard deviation.}
\label{tab:exh-hit-top1}
\begin{tabular}{lcc}
\toprule
Target & Top-1 DS (ex$=1$) & Top-1 DS (ex$=8$) \\
\midrule
PARP1 & $12.490 \pm 0.532$ & $12.830 \pm 0.754$ \\
FA7   & $10.110 \pm 0.367$ & $10.090 \pm 0.335$ \\
5HT1B & $11.780 \pm 0.461$ & $12.240 \pm 0.744$ \\
\bottomrule
\end{tabular}
\end{table}

\begin{table}[htbp]
\centering
\caption{Lead optimization top-1 docking scores for ChemLlama at QuickVina exhaustiveness $1$ versus $8$ ($\delta=0.6$). Entries are mean $\pm$ standard deviation.}
\label{tab:exh-lead-top1}
\begin{tabular}{lcc}
\toprule
Task & Top-1 DS (ex$=1$) & Top-1 DS (ex$=8$) \\
\midrule
lead.5ht1b\_06\_0 & $11.667 \pm 0.153$ & $12.300 \pm 0.900$ \\
lead.5ht1b\_06\_1 & $10.667 \pm 0.764$ & $11.300 \pm 0.700$ \\
lead.fa7\_06\_0   & $9.000 \pm 0.200$  & $9.133 \pm 0.231$ \\
lead.fa7\_06\_1   & $8.367 \pm 0.379$  & $8.333 \pm 0.404$ \\
lead.parp1\_06\_0 & $10.933 \pm 0.577$ & $10.733 \pm 0.153$ \\
lead.parp1\_06\_1 & $12.267 \pm 0.231$ & $12.800 \pm 0.200$ \\
\bottomrule
\end{tabular}
\end{table}

Higher exhaustiveness mostly refines scores, and exhaustiveness $1$ still appears to capture the relative ranking of strong candidates. On a single CPU core, docking takes about $0.05$\,s/molecule at exhaustiveness $1$ versus about $0.25$\,s/molecule at exhaustiveness $8$. Given the consistent improvements at higher exhaustiveness, using exhaustiveness $8$ is a reasonable option when extra compute is available.

\subsection{Adequacy of QuickVina relative to AutoDock Vina}
\label{app:quickvina-vina-correlation}

To check QuickVina, we compare it with AutoDock Vina (1.2.7), an established docking method~\citep{trott2010autodock,eberhardt2021autodock}. We rescore, with AutoDock Vina, the molecules generated by each method on the hit-generation and lead-optimization tasks. Table~\ref{tab:qv-vina-corr} reports the Pearson correlation of the QuickVina and AutoDock Vina docking scores across all molecules generated during optimization (\emph{all}) and across successful molecules (\emph{successful}), aggregated over the target proteins.

\begin{table}[htbp]
\centering
\caption{Pearson correlation between QuickVina and AutoDock Vina docking scores for all generated molecules and for successful molecules, by method and task category.}
\label{tab:qv-vina-corr}
\begin{tabular}{lcccc}
\toprule
Method & Hit all & Hit successful & Lead all & Lead successful \\
\midrule
ChemLlama   & $0.827$ & $0.845$ & $0.676$ & $0.886$ \\
Saturn      & $0.962$ & $0.928$ & $0.882$ & $0.886$ \\
Genetic-guided GFN & $0.972$ & $0.924$ & $0.959$ & $0.885$ \\
GenMol      & $0.922$ & $0.913$ & $0.875$ & $0.680$ \\
\bottomrule
\end{tabular}
\end{table}

We observe that QuickVina and AutoDock Vina are well correlated across methods and molecule sets (\emph{all} and \emph{successful}). These correlation values are also consistent with previously computed correlations between QuickVina and AutoDock Vina on non-AI-optimized molecules~\citep{alhossary2015quickvina}, which was $0.967$. Molecules that score highly under QuickVina also score highly under AutoDock Vina, so the optimization is not fitting idiosyncrasies of QuickVina alone. QuickVina is also considerably faster, which is why the benchmark uses it.

\subsection{Physical plausibility of high-scoring poses}
\label{app:posebusters}

To address whether high-scoring molecules exploit docking artifacts, we additionally tested the highest-scoring molecules across all four methods ($1{,}335$ molecules in total), together with their docked poses, using PoseBusters~\citep{buttenschoen2024posebusters}.

For each molecule, we regenerated a local QuickVina docking pose on the corresponding receptor, preprocessed the resulting complex by removing and re-adding hydrogens to avoid PDBQT conversion artifacts, and evaluated the docked pose with PoseBusters in docked-ligand mode.

Across all methods and tasks, $55.7\%$ of molecules passed all PoseBusters checks. Pass rates were higher for the hit ($59.2\%$) and lead ($56.3\%$) evaluations, and lower for the specificity evaluation ($42.9\%$). As shown in Table~\ref{tab:posebusters-pass}, per-method pass rates range from $50.8\%$ to $60.6\%$, with ChemLlama the lowest.

\begin{table}[htbp]
\centering
\caption{PoseBusters pass rate for the highest-scoring molecules of each method. A molecule passes only if it satisfies every PoseBusters check.}
\label{tab:posebusters-pass}
\begin{tabular}{lc}
\toprule
Method & Pass rate \\
\midrule
Genetic-guided GFN & $60.6\%$ \\
GenMol      & $57.7\%$ \\
Saturn      & $55.0\%$ \\
ChemLlama   & $50.8\%$ \\
\bottomrule
\end{tabular}
\end{table}

We also redocked the $308$ ligands from the PoseBusters benchmark with QuickVina, using the crystal protein and the same PoseBusters protocol as above. $77.3\%$ ($238/308$) passed all checks, and $78.3\%$ of the $304$ successfully docked ligands passed. Under this matched docking and PoseBusters stack, our top-scoring generated molecules pass at a lower rate ($55.7\%$).

\section{Experimental Methodology and Community Resources}
\label{app:methodology}
\label{app:validation-test-split}

This appendix provides hyperparameter sweep procedures, task-specific thresholds, and community resources. The validation/test split, task overview (Figure~\ref{fig:all_tasks}), and model selection protocol are described in Section~\ref{sec:val-test-protocol}.

\subsection{Hyperparameter Sweep Details}
We ran independent hyperparameter sweeps on every validation task for Saturn, GenMol, Genetic-guided GFN, and ChemLlama. Transfer matrices, the category-specific and all-but-me selection rules, and test evaluation follow Sections~\ref{sec:analysis} and~\ref{sec:experiments}. All tuning was confined to validation tasks. Test reporting uses category-specific selection only, with one evaluation per method. Complete search spaces and per-method training details are in Appendix~\ref{app:methods}.

\subsection{Task-Specific Thresholds}
\label{app:task-thresholds}

\subsubsection{Hit Generation Thresholds}
For hit generation tasks, the target-specific docking score thresholds $s_{\min}^{(t)}$ are:
\[
    s_{\min}^{(\mathrm{PARP1})}=10.0,\quad
    s_{\min}^{(\mathrm{FA7})}=8.50,\quad
    s_{\min}^{(\mathrm{5HT1B})}=8.7845,\quad
    s_{\min}^{(\mathrm{BRAF})}=10.30,\quad
    s_{\min}^{(\mathrm{JAK2})}=9.10.
\]

\subsubsection{Specificity Tasks: Reference Docking Score and Property Filters}
Let $s_{\text{ref}}^{(\text{spec})} := s^{(\text{TYK2 JH2})}(\text{Deucravacitinib})$ denote the QuickVina docking score of the reference inhibitor Deucravacitinib (BMS-986165) against TYK2 JH2 under the protocol in Appendix~\ref{sec:docking-details}. We use
\[
    s_{\text{ref}}^{(\text{spec})} = 10.67.
\]
Molecules must satisfy $s^{(\text{TYK2 JH2})}(x)\geq s_{\text{ref}}^{(\text{spec})}$ to count as hits for the specificity objective.

\subsection{Compute Resources}
\label{app:compute-resources}

We have utilized one DGXH100 box, containing 8 NVIDIA H100 GPUs with 80Gb of memory for running hyperparameter sweeps for the different methods and 128 CPU core-processors for deploying and running the QuickVina simulations. Additional computational resources provided through the NVIDIA Academic Grant Program supported follow-up evaluations, ablation studies, and experimental validation conducted during the NeurIPS 2026 rebuttal.

\subsection{Code and Community Contributions}
We invite the community to contribute novel methods and to extend the benchmark with additional objectives. Benchmark wrappers, evaluation code, and baselines accompany this submission. An extended public release with additional baselines and task definitions is available as PMO-Dock: \url{https://github.com/YerevaNN/PMO-Dock}.


\end{document}